\documentclass[journal]{IEEEtran}
\usepackage{times}
\usepackage{epsfig}
\usepackage{graphicx}
\usepackage{amsmath}
\usepackage{amssymb}
\usepackage[utf8x]{inputenc}
\usepackage[style=english]{csquotes}

\usepackage{float}
\usepackage{tabularx}
\usepackage{relsize}
\usepackage{multirow} 
\usepackage{subcaption}
\usepackage{hyperref}

\usepackage[english]{babel}

\usepackage[colorinlistoftodos]{todonotes}
\usepackage{mathtools}
\usepackage{tikz}
\usepackage{relsize}
\usepackage{cleveref}

\def\compmult{\odot}
\def\widthsign{W}
\newcommand{\colj}[1]{#1}
\def\checkmark{\tikz\fill[scale=0.4](0,.35) -- (.25,0) -- (1,.7) -- (.25,.15) -- cycle;}

\graphicspath{{./figures/}}

\newcommand{\reftable}[1]{Table~\ref{#1}}

\newcommand{\figwidth}{\columnwidth}

\newcommand{\ignore}[1]{}

\def\matmult{}
\def\compmult{\odot}

\ifCLASSINFOpdf
\else
\fi
\begin{document}
%
\title{Robust Spatial Filtering with Graph Convolutional Neural Networks}
%
%

\author{Felipe~Petroski~Such*,~\IEEEmembership{Student Member,~IEEE,}
		Shagan~Sah*,~\IEEEmembership{Student Member,~IEEE,}
        Miguel~Dominguez,~\IEEEmembership{Student Member,~IEEE,}
        Suhas~Pillai, 
        Chao~Zhang, Andrew~Michael, Nathan~D.~Cahill,~\IEEEmembership{Senior~Member,~IEEE},
        and~Raymond~Ptucha,~\IEEEmembership{Senior~Member,~IEEE}
        
        *equal contributions}

\twocolumn[
  \begin{@twocolumnfalse}
    This paper is a preprint (IEEE “accepted” status). IEEE copyright notice. © 2017 IEEE. Personal use of this material is permitted. Permission from IEEE must be obtained for all other uses, in any current or future media, including reprinting/republishing this material for advertising or promotional purposes, creating new collective works, for resale or redistribution to servers or lists, or reuse of any copyrighted component of this work in other works. 

Felipe Petroski Such, Shagan Sah, Miguel Dominguez, Suhas Pillai, Chao Zhang, Andrew Michael, Nathan D. Cahill, Raymond Ptucha, Robust Spatial FIltering with Graph Convolutional Neural Networks. Journal of Selected Topics of Signal Processing, Volume PP, Issue 99, 2017
\hyperlink{https://doi.org/10.1109/JSTSP.2017.2726981}{https://doi.org/10.1109/JSTSP.2017.2726981}
  \end{@twocolumnfalse}
]

\clearpage

\maketitle

\begin{abstract}
	Convolutional Neural Networks (CNNs) have recently led to incredible breakthroughs on a variety of pattern recognition problems. Banks of finite impulse response filters are learned on a hierarchy of layers, each contributing more abstract information than the previous layer. The simplicity and elegance of the convolutional filtering process makes them perfect for structured problems such as image, video, or voice, where vertices are homogeneous in the sense of number, location, and strength of neighbors.  The vast majority of classification problems, for example in the pharmaceutical, homeland security, and financial domains are unstructured.  As these problems are formulated into unstructured graphs, the heterogeneity of these problems, such as number of vertices, number of connections per vertex, and edge strength, cannot be tackled with standard convolutional techniques.  We propose a novel neural learning framework that is capable of handling both homogeneous and heterogeneous data, while retaining the benefits of traditional CNN successes. 
   
Recently, researchers have proposed variations of CNNs that can handle graph data. In an effort to create learnable filter banks of graphs, these methods either induce constraints on the data or require preprocessing. As opposed to spectral methods, our framework, which we term Graph-CNNs, defines filters as polynomials of functions of the graph adjacency matrix. Graph-CNNs can handle both heterogeneous and homogeneous graph data, including graphs having entirely different vertex or edge sets. We perform experiments to validate the applicability of Graph-CNNs to a variety of structured and unstructured classification problems and demonstrate state-of-the-art results on document and molecule classification problems.	
\end{abstract}


%
\IEEEpeerreviewmaketitle

\section{Introduction}
Most naturally occurring problems can be described with an underlying graph structure. Functional MRIs, molecules, document databases, social networks, and 3D meshes in computer graphics can all be described by vertices connected by edges. For example, friends are connected through relationships, atoms are connected through bonds, and documents are connected by citations. Making inferences about these graphs and their elements is an active area of research. 


Convolutional Neural Networks (CNNs) have forever changed the pattern recognition landscape with breakthrough results on image classification \cite{krizhevsky2012imagenet, simonyan2014very, Szegedy_2015_CVPR, He2015}, object detection \cite{renNIPS15fasterrcnn, redmon2015you}, and speech recognition \cite{abdelhamid2014asr}. It is natural to want to apply CNN methods to graph data to learn useful features. Graph problems are challenging because graph data does not have the gridded array structure that image, video, and signal data has. Each vertex (e.g. pixel) in gridded structures has the same number of neighbors and the same relationships to a neighbor in a given direction. Non-gridded graphs do not have these limitations. A non-gridded graph can vary in the number of neighbors from vertex to vertex, and there is not necessarily a geometrical interpretation for any given connection between two vertices. 



\begin{figure}[H]
  \centering
    \includegraphics[width=5.5cm]{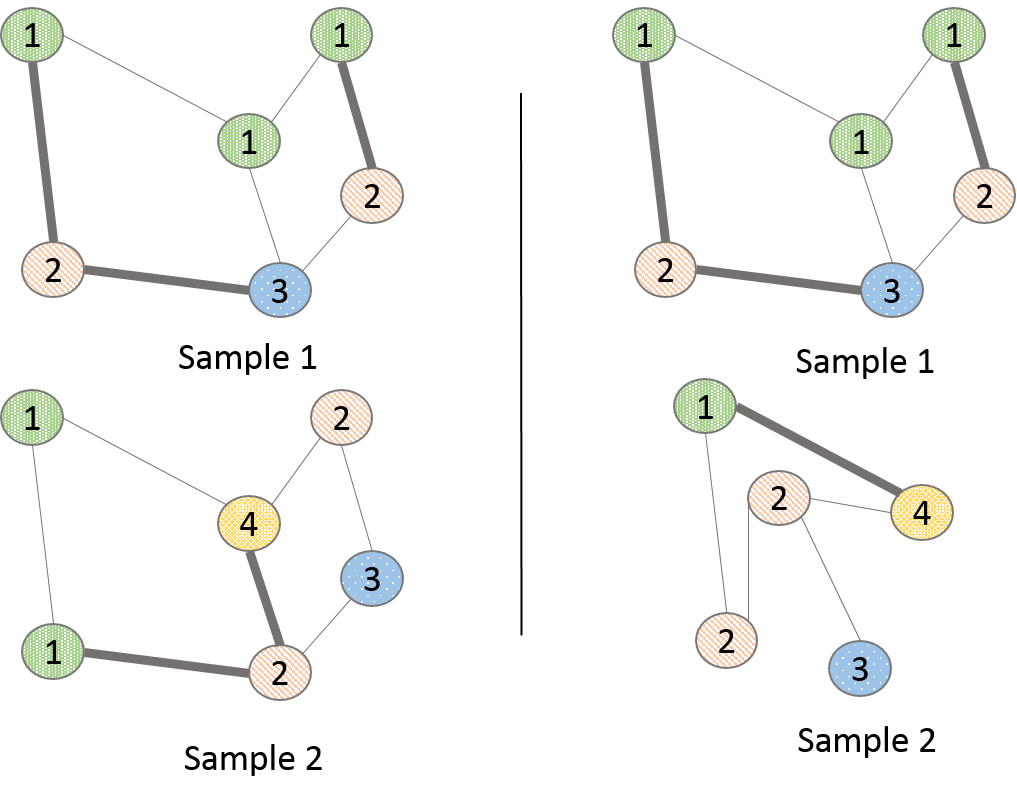}
  \caption{Two types of graph datasets. \textbf{Left:} Homogeneous datasets. All samples in a homogeneous graph data have identical graph structure, but different vertex values or \enquote{signals}. \textbf{Right:} Heterogeneous graph samples.  Heterogeneous graph samples can vary in number of vertices, structure of edge connections, and in the vertex values.}
  \label{fig:graph}
\end{figure}

Filters are elegantly posed as spectral multipliers in a Fourier domain. Pooling can be modeled as spectral graph clustering operations as shown in \cite{henaff2015deep}. Using these filter and pooling operations, numerous studies have applied CNNs on graphs \cite{bruna2013spectral, henaff2015deep, kipf2016semisupervised, ktena2017distance}.  The downside of this approach is that the graphs that are processed by these models are required to be \textit{homogeneous}. This means that each graph sample is required to have the same number of vertices and edge connections, as in Figure \ref{fig:graph}-Left.  The samples can only differ in the \enquote{signal}, that is, the vertex values in the graph. This is because the Fourier domain is unique for each graph. Spectral filters for one graph may not provide the same filtering behavior for another graph. \textit{Heterogeneous} graphs (as in Figure \ref{fig:graph}-Right), which can vary in the number of vertices and the distribution of edge connections, cannot be processed with these models. 

Defferrard et al. \cite{Defferrard2016} introduced graph methods which are spectrally defined, but are implemented spatially with recursive polynomials on the graph Laplacian. This enables spectrally motivated approaches to handle heterogeneous graphs. For example, \cite{ktena2017distance} and \cite{kipf2016semisupervised} use \cite{Defferrard2016} to measure the similarity between functional brain graphs and document classification respectively.  

Sandryhaila, et al. \cite{sandryhaila2013discrete} has shown that a shift-invariant convolution filter can be represented as a polynomial of adjacency matrices. We note the adjacency polynomial \enquote{translation} is an isotropic diffusion from the current vertex to vertices farther away. Like \cite{Defferrard2016}, weights are shared among a graph regardless of (any heterogeneous) structure. Each weight is used by all vertices of a given distance from the current vertex.  We define filters as polynomials of functions of the graph adjacency matrix to define a useful spatial Graph-Convolutional Neural Network.  Like \cite{sandryhaila2013discrete,Defferrard2016}, our work allows filters to be applied to heterogeneous graphs without going into the spectral domain. We also exploit structure in certain graphs (such as images or 3D meshes) to provide anisotropic filtering that varies based on angle. Filters are learned directly from graph adjacency matrices and vertex features in the spatial domain. The code is downloadable from \cite{Felipe2017}.

The contributions of this research are as follows:
\begin{itemize}


\item We introduce the concept of vertex filters on graphs. Vertex filters simultaneously learn properties from both graph vertices and edges. The technique also enables learning from multiple adjacency matrices with individual edge features or graphs.

\item We provide a supervised graph embed pooling operation to learn a pooling transformation for heterogeneous graph data.

\item We do extensive experiments with multiple graph datasets- brain fMRI, chemical compounds, 3D face and document citations, to show the applicability of our model to varying graph structures. We also show that our graph formulation performs similarly to a classical CNNs on CIFAR-10 and ImageNet datasets.


\end{itemize}

\begin{figure*}[!ht]
  \centering
    \includegraphics[width=\textwidth]{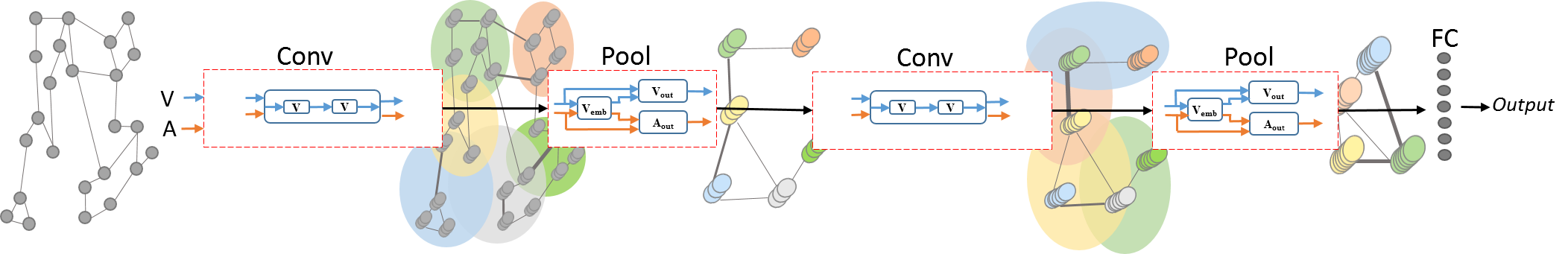}
  \caption{General vertex-edge domain Graph-CNN architecture. Convolution and pooling layers are cascaded into a deep network. FC are fully-connected layers for graph classification. $V$ is vertex set and $A$ is adjacency matrix that define a graph.}
  \label{architecture}
\end{figure*}

This work is a step towards creating a one-to-one mapping between deep convolutional neural networks for signals and images and one for graphs as represented in Figure \ref{architecture}. Further, the presented methods can be applied to graphs that are homogeneous or heterogeneous in nature.

The paper is organized as follows: Section II outlines related work. Section III describes the proposed Graph-CNN model in detail including different filtering techniques. Section IV details the numerical experiments on graph data- images, brain imaging, facial expression recognition, document classification, and chemical compounds. Section V discusses the computational complexity. Section VI contains concluding remarks.

\section{Related work}

In general graph data is encoded by the tuple $\boldsymbol{G}=(\boldsymbol{V},\boldsymbol{A})$. $\boldsymbol{V}\in\mathbb{R}^{N\times C}$ is the vertex data or graph signal, where $N$ vertices each contain $C$ vertex features. $\boldsymbol{A}\in\mathbb{R}^{N\times N}$ is the adjacency matrix, which encodes the connections between vertices. The adjacency matrix entries can be defined as in (\ref{equ:adjentry}).

\begin{equation}
\label{equ:adjentry}
a_{ij} = \begin{cases}
    w_{ij}& \text{if there is an edge between } i \text{ and } j\\ 
    0              & \text{otherwise}
\end{cases}
\end{equation}
The scalar $w_{ij}$ is a weight that represents some measure of strength of the edge between vertex $i$ and vertex $j$.

There are three general approaches we found to generalizing Deep Neural Networks for graph data: spectral, spatial, and geometric. There is some overlap as elements of spectral graph theory are frequently used throughout the literature.
\subsection{Spectral Approaches}
Spectral approaches exploit spectral graph theory. These works filter in a spectral domain by constructing an analogue to the Discrete Fourier Transform (DFT), which is based on the eigenvector decomposition of the Graph Laplacian.
The Graph Laplacian is shown in (\ref{eqn:laplacian}) and the normalized Graph Laplacian is shown in (\ref{eqn:normlaplacian}). $\boldsymbol{A}$ is the adjacency matrix, $\boldsymbol{D}$ is the diagonal degree matrix, whose entries are the row-wise sums of $\boldsymbol{A}$, and $\boldsymbol{I}$ is the identity matrix.
\begin{equation}
\label{eqn:laplacian}
{\boldsymbol{L}= {\boldsymbol{D}}-\boldsymbol{A}}
\end{equation}
\begin{equation}
\label{eqn:normlaplacian}
{\boldsymbol{L}= {\boldsymbol{I} - \boldsymbol{D}^{-1/2} \boldsymbol{A} \boldsymbol{D}^{-1/2}}}
\end{equation}
$\boldsymbol{L}$ can be used to compute an eigenbasis $\boldsymbol{U}$ that represents an analogue to the DFT matrix. Then a graph signal $\boldsymbol{x}$, which contains the vertex values of the graph, can be filtered spectrally by transforming $\boldsymbol{x}$ into the spectral domain and multiplying each frequency by a filter $\boldsymbol{h}$, as in (\ref{eqn:spectralfilter}) ($\odot$ is the elementwise product).

\begin{equation}
\label{eqn:spectralfilter}
\boldsymbol{x} * \boldsymbol{h} = \boldsymbol{U}^T \cdot (\boldsymbol{Ux} \odot \boldsymbol{h})
\end{equation}

Eigenvectors of the Graph Laplacian represent frequency components, similar to rows of the DFT matrix. By transforming graph signals to a spectral domain with the resulting eigenbasis, the graph signal can be multiplied by an array of filter coefficients to perform a filtering operation. Several works propose graph CNN models that are based on this method of filtering  \cite{bruna2013spectral,henaff2015deep,Defferrard2016,edwards2016graphcnn,kipf2016semisupervised}. One of the advantages of this is that these works can leverage a mature body of literature on spectral clustering to propose effective graph-pooling mechanisms (an introduction to spectral clustering can be found in \cite{vonLuxburg2007}). Some also use off-the-shelf software such as Graclus \cite{dhillon2007}.

One initial concern about spectral approaches to graph convolutions is that these filters would not be localized in the spatial domain, meaning that learned weights would not be shared across different locations in the graph. Recent developments have counterintuitively shown that spectral filters can be localized in space. In \cite{bruna2013spectral,henaff2015deep,edwards2016graphcnn}, smooth spectral filters lead to localized filters in the spatial domain, leading to localized filters. In \cite{Shuman2016,Defferrard2016}, it is shown that a $K$-order polynomial formulation of the graph Laplacian perform a $K$-hop filtering operation, despite being a \enquote{spectral} operation. In addition \cite{Defferrard2016} reveals an efficient recursive approximation of this spectral filtering using Chebyshev polynomials. This technique is used for semi-supervised document classification in \cite{kipf2016semisupervised}. Graph signals filtered with this technique are fed through an LSTM in \cite{seo2016graphrecurrent} to model sequences.

One of the major practical limitations when learning filters in the spectral domain is that the eigenbasis that transforms a graph between spatial and spectral domains is unique for each graph. This requires input samples to be homogeneous. The Graph Laplacian eigenbasis needs to be solved separately for each unique graph structure. Most spectral works tend to focus on experiments where there is a single graph structure common across all samples, such as the MNIST image dataset \cite{lecun98}. More recent works based on polynomials of the Laplacian do not have this limitation \cite{kipf2016semisupervised,ktena2017distance,Defferrard2016}. By representing polynomial filters as linear combinations of Chebyshev polynomials, they are able to exploit the Chebyshev polynomial recurrence relation to rapidly apply these filters in the spatial domain (An interesting aspect of this approach is that it is motivated by spectral graph theory, but implemented spatially).

\subsection{Spatial Approaches}
Spatial approaches have an advantage over some spectral approaches in that they do not, by nature,  require a homogeneous graph structure. However, they generally require sophisticated data preprocessing to enable learning. The challenge is in figuring out how to process neighborhoods that are different sizes and structures for each vertex. Diffusion-Convolutional Neural Networks (DCNNs) \cite{atwood2015diffusion} model the graph by encoding layers of matrices that arrange vertex features based on a sequence of hops from different starting vertices. However, due to lack of a vertex pooling or clustering layers, it was not expanded to learn beyond the original level of abstraction. PATCHY-SAN attempts to linearize a graph based on the CNN concept of a receptive field \cite{niepert2016cnn}. \cite{DBLP:journals/corr/DuvenaudMAGHAA15} defines a hashing operation inspired by CNN properties and uses it to learn features on molecular fingerprints. Gated Graph Sequential Neural Networks \cite{ggnn2016} presented a graph Long Short-Term Memory (LSTM) Recurrent Neural Network model applied to program verification and basic logical reasoning tasks, extending a neural network model that learned a graph encoding from data \cite{scarselli2009graph}. DeepWalk \cite{perozzi2014deepwalk} learns latent representations of graph vertices by using random walks to extract local information which encodes structural regularities in social networks.

Our work is also a spatial approach. Like the spectral approaches, our work is inspired by discrete graph signal processing theory.  Specifically, we benefit from the spatial filtering theory proposed in \cite{sandryhaila2013discrete}. Where we differ from current spatial approaches is that we require less preprocessing to format the graph structure. We learn convolutional filters directly on the adjacency matrices and vertex features that naturally represent graph data.

\subsection{Geometric Approaches}
Researchers from the 3D shape analysis community have also been working on the problem of signal processing on heterogeneous data structures. Several recent works have tried to develop filters and CNNs for manifolds, oftentimes for the purpose of point correspondence on 3D meshes. One popular approach is to map individual patches of these manifolds to an alternative representation that is more amenable to filtering. The idea is that filtered patches of similar objects (e.g. fingertips) should have similar feature representations between meshes that only differ in deformation. Some example representations are 2D polar coordinate representations that are filtered with a polar convolution\cite{masci2015geodesic}, local windowed spectral representations \cite{boscaini2016classdesc}, anisotropic variants of heat kernel diffusion filters and spectral filters \cite{boscaini2016anisotropicdesc,boscaini2016anisotropiccnn}, and learned Gaussian Mixture-Model kernels \cite{monti2016geometric}. A survey of this approach is provided in \cite{bronstein2016geometricdeep}. The Gaussian Mixture Model method was generalized to graph data and applied to MNIST classification and document classification \cite{monti2016geometric}.

\section{The Graph-CNN Model}
Our work uses the standard definition of graphs as described above. We place no constraints on $\boldsymbol{A}$ beyond what is described in (\ref{equ:adjentry}). We allow the graph to be directed ($a_{ij}$ does not necessarily equal $a_{ji}$), have reflexive connections ($a_{ii}$ is a valid connection), and different graph samples to be heterogeneous: each sample may have different numbers of vertices and differing graph structures. This latter feature is an improvement over many spectral methods, where filters are learned on a particular \enquote{homogeneous} arrangement of vertices and edges.

For our work, we also define an adjacency tensor $\boldsymbol{\mathcal{A}}\in\mathbb{R}^{N\times N\times L}$, which is a stack of $L$ adjacency matrices $(\boldsymbol{I}, \boldsymbol{A}_1, \boldsymbol{A}_2 \ldots \boldsymbol{A}_{L-1})$, where $\boldsymbol{I}$ is the identity matrix (only reflexive connections). This allows us to encode multiple edge features or to partition edges based on a hand-picked structure.

Throughout this work we utilize the notation defined in \ref{tab:notation}.

\begin{table}[!ht]
\centering
\caption{Graph-CNN notation.}
\label{tab:notation}
\setlength\extrarowheight{3pt}
\begin{tabularx}{7.3cm}{|c|c|p{4cm}|}
\hline
 & Dimensions & Description \\ \hline
$N$ & $\mathbb{R}$ & \# of vertices \\ \hline
$C$ & $\mathbb{R}$ & \# of vertex features \\ \hline
$L$ & $\mathbb{R}$ & \# of edge features \\ \hline
$F$ & $\mathbb{R}$ & \# of filters \\ \hline
$\boldsymbol{V}$ & $\mathbb{R}^{N\times C}$ & Vertex matrix \\ \hline
$\boldsymbol{A}$ & $\mathbb{R}^{N\times N}$ & Adjacency matrix\\ \hline
$\boldsymbol{\mathcal{A}}$ & $\mathbb{R}^{N\times N\times L}$ & Adjacency tensor (multiple $\boldsymbol{A}$s) \\ \hline
$\boldsymbol{H}$ & $\mathbb{R}^{N\times N\times C\times F}$ & Graph filter \\ \hline
\end{tabularx}
\end{table}

\subsection{Graph Filters} \label{subsection:ModifiedSpatialDomainFiltering}

Sandryhaila, et al. \cite{sandryhaila2013discrete} recently defined a spatial-domain convolution for graphs, shown in (\ref{equ:spatial_filter}).
\begin{equation}
\label{equ:spatial_filter}
\boldsymbol{H} = h_0\boldsymbol{I} + h_1\boldsymbol{A} + h_2\boldsymbol{A}^2 + ... + h_k \boldsymbol{A}^k, \boldsymbol{H}\in\mathbb{R}^{N\times N}
\end{equation}
The filter is defined as the $k$th-degree polynomial of the graph's adjacency matrix. Each exponent in the polynomial encodes the number of hops from a given vertex that are being multiplied by the given filter tap. $\boldsymbol{A}^1$ (or $\boldsymbol{A}$) represents the one-hop neighbors of the given vertex. $\boldsymbol{A}^2$, the square of the adjacency matrix, represents the two-hop neighbors, and so on. $\boldsymbol{I}$ represents the 0-hop or the vertex being processed. The scalar coefficients $h_0$, $h_1$,...,$h_k$ control the contribution of the neighbors of a vertex during the convolution operation.

To convolve the vertices $\boldsymbol{V}$ with the filter $\boldsymbol{H}$ is a matrix multiplication, $\boldsymbol{V_{out}}=\boldsymbol{HV_{in}}$ where $\boldsymbol{V_{in}},\boldsymbol{V_{out}}\in\mathbb{R}^{N}$.




\subsection{Graph-CNN}\label{subsection:GraphCNN}
To adapt this filtering operation into a convolution operation fit for our Graph-CNN, we need to take into account the desire to process multiple filters and multiple adjacency matrices per sample. We also want to use the intuition from VGGNet \cite{simonyan2014very} that learning cascades of small filters can effectively capture the receptive field of a single large filter. To that end we approximate (\ref{equ:spatial_filter}) as a linear equation in (\ref{equ:linearfilter}).
\begin{equation}
\label{equ:linearfilter}
\boldsymbol{H} \approx h_0\boldsymbol{I} + h_1\boldsymbol{A}
\end{equation}
This equation is cascaded over multiple layers in a neural network, thereby achieving the receptive field of \ref{equ:spatial_filter} but with nonlinearities at every step. A comparable linear approximation for a spectral approach was used by Kipf, et al. \cite{kipf2016semisupervised}. Their approach similarly cascaded these linear filters to approximate the $K$-hop filter formed by the polynomial of the Laplacian.

\begin{figure}[H]
  \centering
    \includegraphics[width=\figwidth]{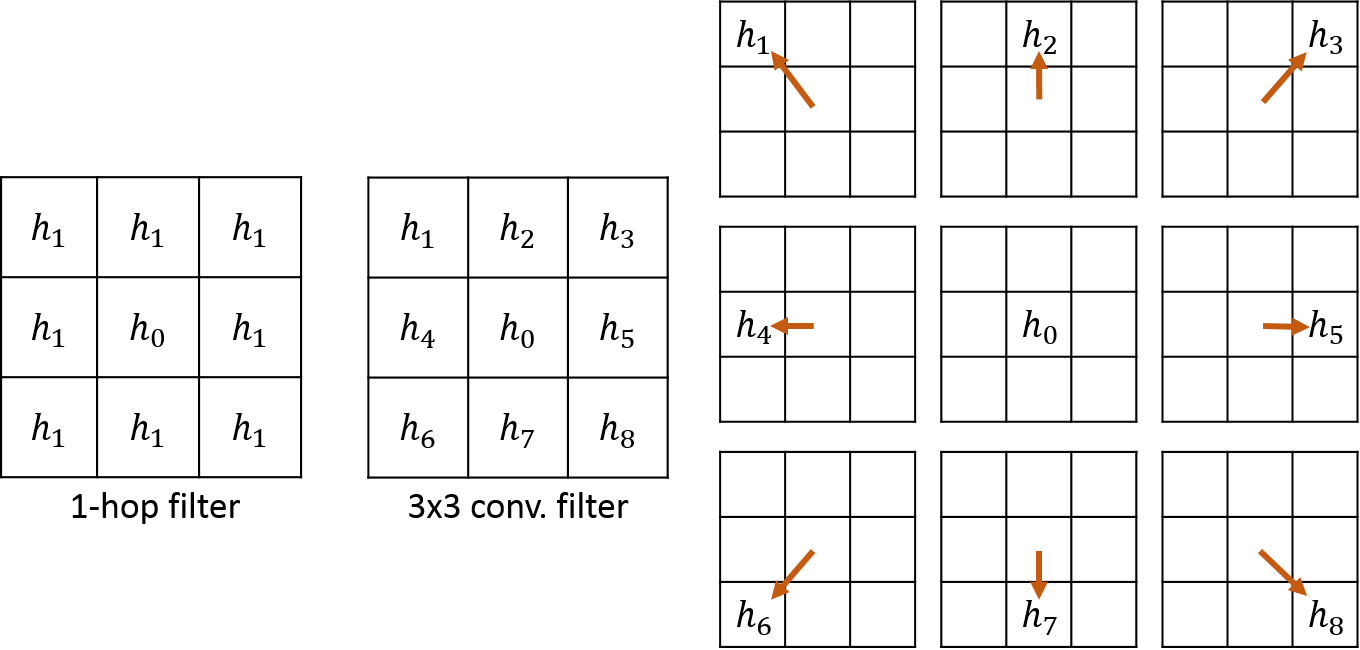}
  \caption{\textbf{Left:} Learnable parameters in 1-hop graph filters. \textbf{Center:} Classical $3\times3$ convolution filters. \textbf{Right:} Illustration of eight different edge connections combined to form a $3\times3$ filter.}
  \label{fig:filter}
\end{figure}

We want to scale this operation up to use our adjacency tensor $\boldsymbol{\mathcal{A}}$. This construct contains multiple adjacency matrices in a sample. The first slice of this tensor is $\boldsymbol{A}_1$, the second slice is $\boldsymbol{A}_2$ and so on. Each slice $\boldsymbol{A}_\ell$ encodes a particular edge feature for the graph in an adjacency matrix. We define a linear filter as a convex combination of each adjacency matrix as in (\ref{equ:linearfiltermultiple}).
\begin{equation}
\label{equ:linearfiltermultiple}
\boldsymbol{H} \approx h_0\boldsymbol{I} + h_1\boldsymbol{A}_1 + h_2\boldsymbol{A}_2 + \ldots h_{L-1}\boldsymbol{A}_{L-1}
\end{equation}
This can be written compactly as in (\ref{equ:linearfiltermultipletensor}).

\begin{equation}
\label{equ:linearfiltermultipletensor}
\boldsymbol{H} \approx \sum_{\ell=1}^{L} h_\ell\boldsymbol{A}_\ell
\end{equation}
One motivation for these multiple adjacency matrices is to encode multiple edge features, one feature in each $\boldsymbol{A}$. Another is to partition edges in a single $\boldsymbol{A}$ into multiple matrices to impart a sense of direction. Figure \ref{fig:filter} illustrates this motivation. The figure on the left is an illustration of the default Graph-CNN linear filter in an image application. A given filter tap is applied to all vertices of a given distance, isotropically. In this case, $h_0$ is applied to the 0-hop vertex and $h_1$ is applied to all adjacent vertices. If another border of pixels surrounded this figure, each pixel in that border would be multiplied by a filter tap $h_2$.

If however, the adjacency matrix was partitioned into nine adjacency matrices, each one representing a different relative connection to a given vertex (upper-left, right, down, lower-right, and so on), then there would be a unique filter tap for each direction, including the 0-hop self-connection (Figure \ref{fig:filter}, right). This results in an anisotropic Graph-CNN filter. This is equivalent to a $3\times 3$ FIR filter in conventional CNN applications (Figure \ref{fig:filter}, center). The Appendix furnishes a proof of this that explains the model in more detail. For graph datasets such as images and 3D meshes that have exploitable structure, this method can multiply parameters and increase the modeling capability of the network.

So far we have described filters for a single vertex feature. To have a set of filter coefficients for multiple vertex features, each $h_\ell$ needs to be in $\mathbb{R}^{C}$, leading $\boldsymbol{H}$ to be in $\mathbb{R}^{N\times N\times C}$. This means $\boldsymbol{H}$ is a stack of $N\times N$ filter matrices indexed by the vertex feature they filter. Equation (\ref{equ:linearfiltermultipletensor}) can be modified as in (\ref{equ:linearfiltermultipletensormultiplefeature}) to illustrate this new operation.
\begin{equation}
\label{equ:linearfiltermultipletensormultiplefeature}
\boldsymbol{H}^{(c)} \approx \sum_{\ell=1}^{L} h_\ell^{(c)}\boldsymbol{A}_\ell
\end{equation}
In (\ref{equ:linearfiltermultipletensormultiplefeature}), $\boldsymbol{H}^{(c)}$ is an $N\times N$ slice of $\boldsymbol{H}$ and $h_\ell^{(c)}$ is a scalar corresponding to a given input feature and a given slice of $\boldsymbol{A}_\ell$. To filter a vertex signal $\boldsymbol{V}_{in}$ using this filter tensor, we perform the operation described in (\ref{equ:onetensorfilter}).
\begin{equation}
\label{equ:onetensorfilter}
\boldsymbol{V}_{out} = \sum_{c=1}^{C}\boldsymbol{H}^{(c)}\boldsymbol{V}_{in}^{(c)} + b
\end{equation}
$\boldsymbol{V}_{in}^{(c)}$ is the column of $\boldsymbol{V}_{in}$ that only contains vertex feature $c$. We also add a bias $b\in\mathbb{R}$. This results in $\boldsymbol{V}_{out}\in\mathbb{R}^N$. This is analogous to an image with multiple color channels being filtered down to a single grayscale channel in image-based CNNs. Multiple filters can be modeled by adding another dimension to $\boldsymbol{H}$ so that it can become part of $\mathbb{R}^{N\times N\times C\times F}$. Then (\ref{equ:onetensorfilter}) can be repeated, with each filter output representing a single column $\boldsymbol{V}_{out}^{(f)}$ in $\boldsymbol{V}_{out}\in\mathbb{R}^{N\times F}$. Each feature in the output vertices is the output of a single filtering operation across all features in the input vertices. In this case there are also $F$ biases, one for each filter.

\begin{figure}[!ht]
\centering
\begin{subfigure}{.5\figwidth}
  \centering
    \includegraphics[height=1.3\figwidth]{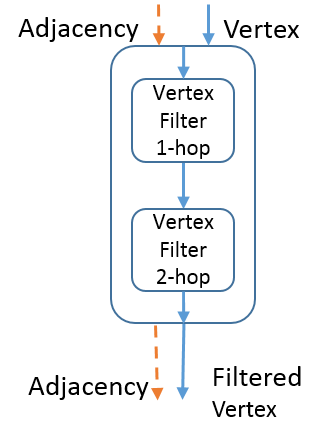}
    \caption{Graph Convolution}
    \label{fig:filter_img}
\end{subfigure}%
\begin{subfigure}{.5\figwidth}
  \centering
    \includegraphics[height=1.3\figwidth]{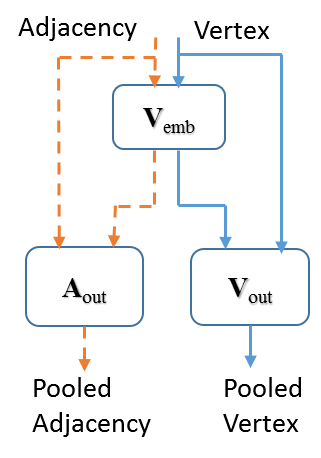}
    \caption{Graph Pooling}
    \label{fig:pool_img}
\end{subfigure}
  \caption{Graph convolution and pooling setting. The convolution operation obtains a filtered representation of the graph after a multi-hop vertex filter. Likewise, a compact representation of the graph after a pooling layer.}
\end{figure}

To be clear, these vertex filters only change the vertex data. The adjacency data is used to help filter the vertices, but remains unchanged by the operation. Figure \ref{fig:filter_img} illustrates this.

\subsection{Initialization}

Like other neural network models, Graph-CNNs require proper weight initialization. A common way of doing so is by initializing the weights with random numbers generated from a Gaussian distribution. This works well with small networks, but deep networks suffer from vanishing (or exploding) gradients. Xavier initialization \cite{GlorotAISTATS2010} addresses this somewhat, by intelligently selecting the Gaussian distribution parameters to mitigate changes in the distribution of the input and output data. Xavier bases the parameters on the number of inputs and number of outputs of each filter. A more complex variant would be required for graph data. This is because the output distribution depends on the weights, the input vertices, and the input adjacency matrix. Batch Normalization \cite{ioffe2015batch} offers a more elegant solution. By normalizing each batch of data at each layer, vanishing and exploding gradients are explicitly prevented.

\subsection{Graph Embed Pooling}
\label{subsec:graphpool}

An important building block of CNNs are pooling layers. Reducing dimensions of the input allows convolution filters to have a larger receptive field and also improves computation performance. One of the most common methods for pooling images is max-pooling. This method selects the top value over a defined region in a sliding window approach.

Pooling methods designed for images are tailored for gridded structures and cannot be applied to general graphs due to their often heterogeneous structure.  As a solution, we introduce a method called \textit{graph embed pooling}.  Graph embed pooling learns a convolutional layer whose output can be treated as an embedding matrix that produces a fixed-size output. To produce a pooled graph reduced to a fixed $N'$ vertices, the learned filter taps from this pooling layer produce an embedding matrix $\boldsymbol{V}_{emb}\in\mathbb{R}^{N\times N'}$. Similar to (\ref{equ:onetensorfilter}), a filter tensor $\boldsymbol{H}_{emb}\in\mathbb{R}^{N\times N\times C\times N'}$ is learned and multiplied by the vertices to produce a filtered output, as in (\ref{eqn:embed}). Equation (\ref{eqn:embed}) could be replaced with any layer that produces a fixed number of features for each vertex.  We use the previously defined 1-hop Graph-CNN filter.
\begin{equation}
\label{eqn:embed}
\boldsymbol{V}_{emb}^{(n')} = \sum_{c=1}^{C}\boldsymbol{H}_{emb}^{(c,n')}\boldsymbol{V}_{in}^{(c)} + b
\end{equation}
Each output in the equation is a column $\boldsymbol{V}_{emb}^{(n')}$ in the output matrix $\boldsymbol{V}_{emb}$, indexed by row $n'\in 1, 2,\ldots N'$. Like the other filter tensors, $\boldsymbol{H}_{emb}$ is produced in an analogous fashion to (\ref{equ:linearfiltermultipletensormultiplefeature}). This means that a $O(N^2CN')$ variable-dimension embedding matrix $\boldsymbol{H}_{emb}$ ($N$ is variable) can be learned with $O(LCN')$ parameters. Equations (\ref{equ:graphembed_v}) and (\ref{equ:graphembed_A}) show how $\boldsymbol{V}_{emb}$ produces the pooled graph data. The embedding values of $\boldsymbol{V}_{emb}$ are normalized using a softmax operation ($\sigma$). Note that in graph embed pooling, both the adjacency matrix and the vertices are transformed, as in Figure \ref{fig:filter_img}.

\begin{equation}
\boldsymbol{V}_{emb}^* = \sigma(\boldsymbol{V}_{emb})
\label{equ:graphembed_norm}
\end{equation}

\begin{equation}
\boldsymbol{V}_{out} = \boldsymbol{V}_{emb}^{*T}\boldsymbol{V}_{in}
\label{equ:graphembed_v}
\end{equation}

\begin{equation}
\boldsymbol{A}_{out} = \boldsymbol{V}_{emb}^{*T}\boldsymbol{A}_{in}\boldsymbol{V}_{emb}^*
\label{equ:graphembed_A}
\end{equation}

There are two advantages to graph embed pooling. First, it flexibly takes input of any cardinality or structure and produces a fixed size output. This output can be of any cardinality $N'$. Second, this pooling is learned, so the output structure is the one that represents a reduced-dimension input structure in at least a locally optimal way, similar to other embedding methods. In our work, we use a special case of graph embed pooling where $N'=1$ to produce a graph representation or Graph Fully-Connected (GFC) vector, an embedding we use in some of our experiments later in the paper.

\begin{figure}[!ht]
\centering
\begin{subfigure}{.33\figwidth}
  \centering
    \includegraphics[height=0.75\figwidth]{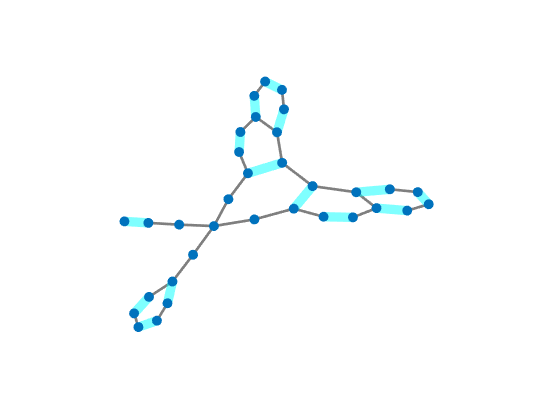}
    \caption{Input graph.}
    \label{fig:pool_left}
\end{subfigure}%
\begin{subfigure}{.32\figwidth}
  \centering
    \includegraphics[height=0.75\figwidth]{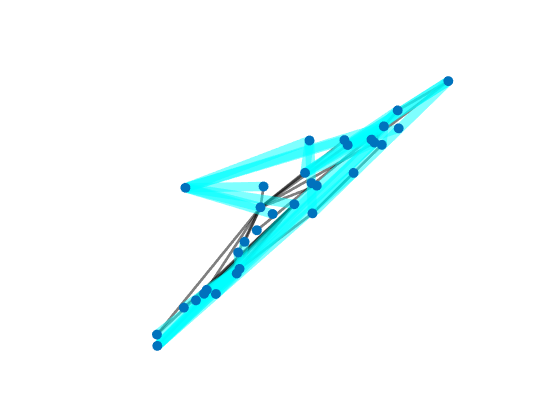}
    \caption{Pool to 32 vertex.}
    \label{fig:pool_center}
\end{subfigure}
\begin{subfigure}{.32\figwidth}
  \centering
    \includegraphics[height=0.75\figwidth]{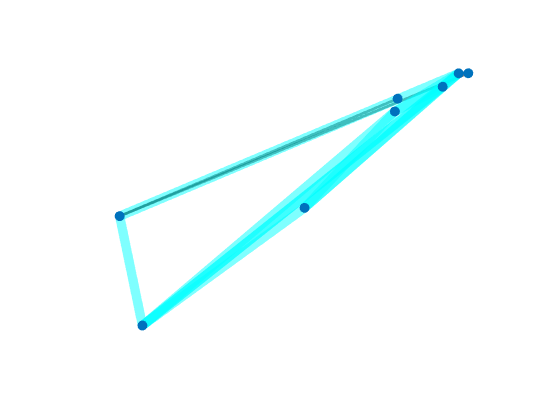}
    \caption{Pool to 8 vertex.}
   \label{fig:pool_right}
\end{subfigure}
  \caption{Graph embed pooling representation as applied to a graph. ($x$, $y$) positions are synthesized for visual purposes. The graph embed pooling learned is applied resulting in (b) and then again in (c). Graphs show top 10\% of edges and are not drawn to scale.}
  \label{fig:pool}
\end{figure}

One particular notion that must be realized is that visually this pooling does not resemble the intuition of average or max pooling in images, where the output signal would seem like a lower-resolution approximation of the input signal. In graph embed pooling, the output vertices are a convex combination of the input vertices, and they are fully connected.  Further, this methodology induces self-connections.  We use the simpler self-connections in \ref{eqn:embed} $-$ \ref{equ:graphembed_A}, because we notice it does not measurably change the resulting performance of the models. Figure \ref{fig:pool} is an illustration of the pooling process, but the geometry of the figure should not be taken literally.

\section{Experiments}

We apply our Graph-CNN model to five different problems.  First, we compare Graph-CNN to traditional CNNs using the CIFAR-10 and ImageNet image classification datasets. Second, we perform gender classification based on Human Connectome Project (HCP) fMRI data.  Third, we classify chemical compounds with the NCI1 and D\&D datasets. Fourth, we classify facial expressions based on the Bosphorus 3D face dataset.  Finally, we evaluate document classification with the Cora document datasets. These problems explore both homogeneous and heterogeneous datasets.

In each investigation, the Graph-CNNs are learned via stochastic gradient descent or Adam optimization \cite{kingma2014adam} using back-propagation. For learning graph filters, the base learning rate is 0.01 and we use a momentum of 0.9 during updates. All architectures use ReLU activation function. Batch Normalization \cite{ioffe2015batch} was also used in all but the last layers of each architecture.

Most of the evaluations attempt graph classification, which attempts to apply a single label to an entire graph. The Document Classification task is a vertex classification task, which means the Graph-CNN attempts to apply a label to each individual test vertex based on the neighboring training vertices.

\subsection{Image Classification}
\begin{figure}[!ht]
  \centering
    \includegraphics[width=2.5cm]{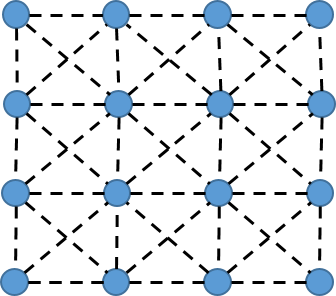}
  \caption{Representation of an image as a gridded graph. Vertices are pixels and lines are edge connections.}  \label{fig:grid}
\end{figure}

For a classification task, an image can be represented as a graph: a two-dimensional rectangular grid as shown in Figure \ref{fig:grid}. We run image classification experiments to give empirical evidence to the earlier claim that classical CNN can be modeled with Graph-CNN's with appropriate adjacency matrices. We use the CIFAR-10 \cite{krizhevsky2009learning} and the ImageNet dataset \cite{russakovsky2015imagenet}.

\subsubsection{CIFAR-10}
Every image in CIFAR is $32 \times 32$ pixels with RGB channels. They can each be represented as a graph with 1024 vertices, where every pixel is a vertex in the graph.  We compare classification methods by learning the same number of parameters. Graph-CNN utilizes the 8-adjacency connection tensor described in Figure \ref{fig:filter}. The CNN$_{3\times3}$ is a standard CNN with $3\times3$ convolution filters. \reftable{cifar_results} shows classification accuracy on the CIFAR-10 dataset. We also compare spectral filters to CNNs and Graph-CNNs in the appendix. All models were trained for 60 epochs.

\begin{table}[!ht]
\centering
\caption{CIFAR-10 image graph classification results. $nF$ is a convolution layer with $n$ filters, $P/2$ is a max-pooling $/2$ layer and FC is a fully connected layer with 128 hidden vertices. All models have a FC-10 layer and a softmax loss layer.}

\label{cifar_results}
\begin{tabular}{|c|c|c|c|}
\hline
\multirow{2}{*}{\#layers} & \multirow{2}{*}{Architecture} & \multicolumn{2}{c|}{Accuracy (\%)} \\ \cline{3-4}
                           &                & Graph-CNN & $\text{CNN}_{3\times3}$   \\ \hline
\multirow{2}{*}{1}         & 32F-FC         & 62.51     & 62.11              \\ \cline{2-4} 
                           & 64F-FC         & 64.12     & 63.4               \\ \hline
\multirow{2}{*}{2}         & 32F-P/2-32F-FC & 66.15     & 67.42              \\ \cline{2-4} 
                           & 32F-P/2-64F-FC & 67.54     & 68.36              \\ \hline
3                          & 3$\times$(32F-P/2)-FC & 68.33      & 68.8              \\ \hline
\end{tabular}
\end{table}


\subsubsection{ImageNet}
\begin{figure}[!ht] 
  \centering
    \includegraphics[width=\figwidth]{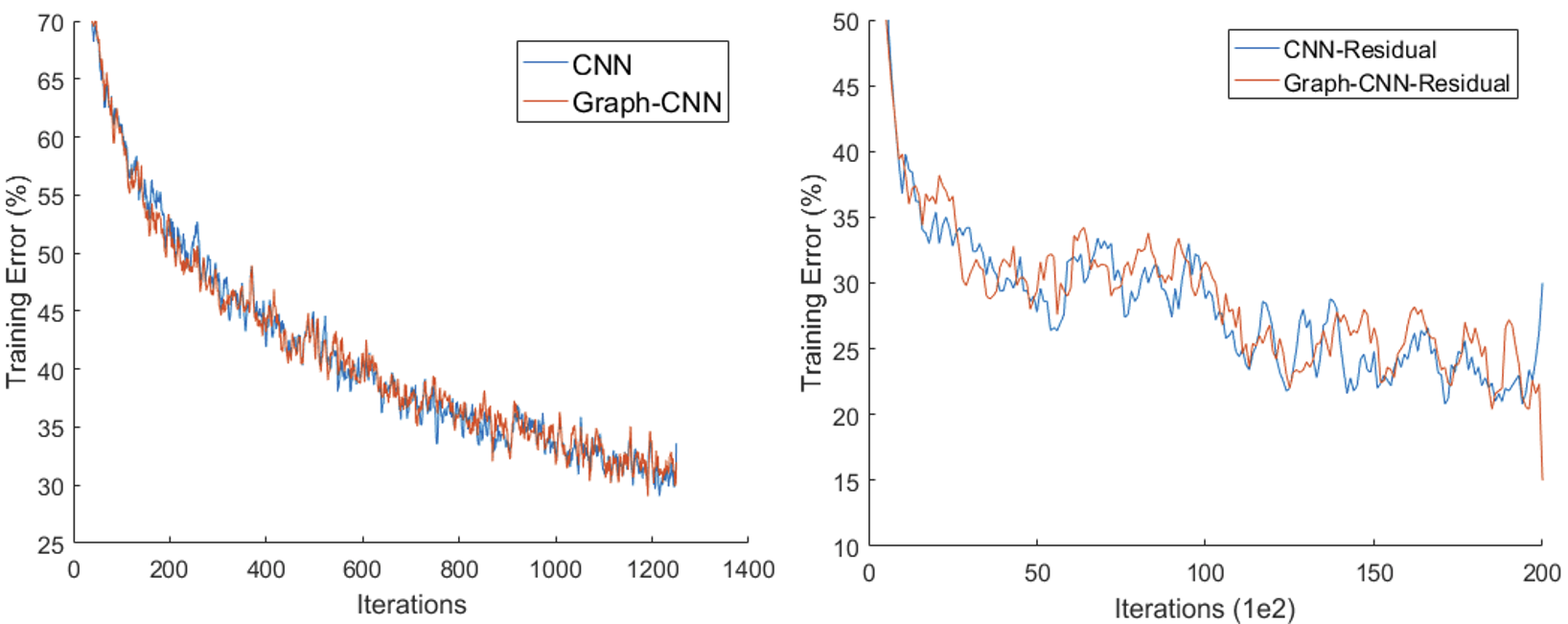}
  \caption{Comparison of training error for Graph-CNN and CNN architectures. \textbf{Left:} CIFAR-10 three layer architecture. \textbf{Right:} ImageNet with ResNet-152 architecture. Both the models show high compatibility between Graph-CNN and CNN.}
  \label{fig:resnet_comp}
\end{figure}

We also evaluate our method on the ImageNet 2012 image classification dataset. We use the ResNet-152 architecture \cite{He2015} to demonstrate the compatibility between Graph-CNN and CNN. We replace the last residual layer (res5) with its Graph-CNN equivalent. We learn for 2e4 iterations and measure top-1 accuracy and use no data augmentation.  With the unmodified ResNet we achieve 70.32\% accuracy. With the ResNet modified with Graph-CNN layers, we achieve 70.02\%.  Though the accuracy we report is less than the actual ResNet performance, there is less than 1\% disparity between the two values. Additionally, the learning curve in Figure \ref{fig:resnet_comp} demonstrate that both CNN and Graph-CNN learn at the same rate.

The CIFAR and ImageNet experiments show that traditional CNN's can be modeled as Graph-CNN's. Since the formulation of the adjacency matrices allow for same number of learnable parameters in the convolution filters, the learning curves for both the datasets indicate high agreement between Graph-CNN and traditional CNN.

\subsection{Human Connectome Project}

The Human Connectome Project (HCP) \cite{hcpOnline} contains resting-state functional Magnetic Resonance Imaging (rsfMRI) data from each of 366 male and 454 female subjects. We seek to investigate whether we can reliably identify the subjects' biological sex from this data.
We use the rsfMRI data preprocessed according to HCP's Minimal Preprocessing Pipeline \cite{glasser2013mpp}. We then subsequently apply denoising to the data according to the FIX protocol \cite{salimi2014adf}. Each FIX-preprocessed rsfMRI is parcellated according to the Automatic Anatomical Labeling (AAL) atlas \cite{tzourio2002aal}, which divides the brain into 90 cortical/subcortical regions and 26 cerebellar/vermis regions. Time series from voxels within each region are averaged to form a representative time series for each region.

We introduce four different weighted adjacency matrices modeling the functional connectivity of each subject based on computing similarities between the representative time series for each region: Fisher $z$-transformed Pearson correlation coefficients (CORR), Fisher $z$-transformed partial correlations (PCORR), and uniform (UNFM) and adaptive (ADPT) versions of the structure-aware affinity inference model for capturing subtle information distributed over discriminative feature subspaces \cite{zhu2014constructing}. These weighted adjacency matrices have the same vertices for each subject but different edge weights.

\begin{figure}[!ht] 
  \centering
    \includegraphics[width=\figwidth]{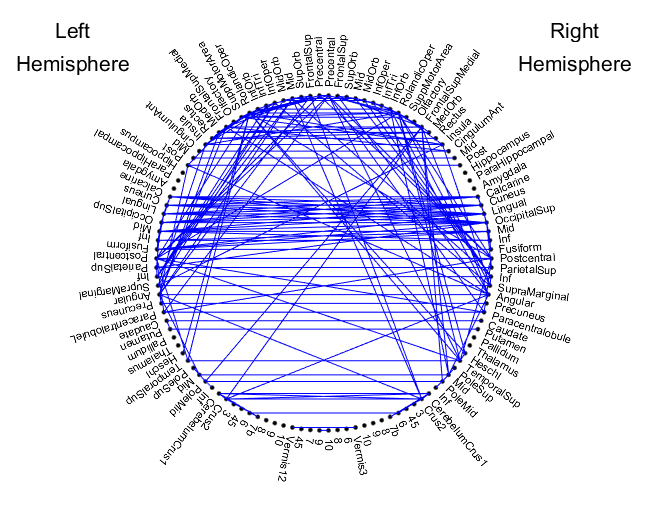}
  \caption{Graph-based representation of AAL atlas regions. Labels indicate the corresponding brain region, and blue lines correspond to the top 3\% most correlated edges (edge weights) in the groupwise average CORR graph (across all male and female subjects).}
\end{figure}

\begin{table}[!ht]
\centering
\caption{Comparison with different input graph matrices. All models have the architecture 2x(32F)-FC.}
\label{mri_1}
\begin{tabular}{|c|c|}
\hline
Input data & Accuracy(\%) \\ \hline
CORR       & $74.51 \pm3.74$   \\ \hline
PCORR      & $74.75 \pm3.30$   \\ \hline
UNFM       & $\textbf{83.78} \pm3.89$   \\ \hline
ADPT       & $78.90 \pm3.18$   \\ \hline
CORR + PCORR + UNFM + ADPT       & $74.26 \pm2.49$   \\ \hline
UNFM + ADPT       & $78.90 \pm3.18$   \\ \hline
\end{tabular}
\end{table}

Experiments were run with various types of similarity matrices. These are listed in \reftable{mri_1}. It also reports results with combinations of the adjacency matrices. Multiple adjacency matrices are treated as edge features while defining the convolution filters. Since the number of training samples is less, complex structures formed through multiple adjacency matrices degrade the performance.
The uniform representation of the similarity are listed in \reftable{mri} for this classification task. We observe that the best-performing model is the 2-layer Graph-CNN model with UNFM input matrices. We report a maximum accuracy of 83.78\% in classifying male vs female using just fMRI graphs.  
These findings reveal that differences in intrinsic connectivity as measured with rs-fMRI exist between subjects. The Graph-CNN filters are capable of detecting and utilizing these differences for classification and gender prediction.

With increased number of imaging-based clinical studies on various diseases such as autism, attention-deficit hyperactivity disorder, etc. \cite{vergun2013characterizing}, the Graph-CNN seems a promising approach for distinguishing disease states from healthy brains on the basis of measurable differences in spontaneous activity. As the amount of available rs-fMRI data increases, learning based methods will be able to extract more meaningful information which can be used in complement with human clinical diagnoses to improve overall efficacy.

\begin{table}[!ht]
\centering
\caption{Male-female classification using the UNFM matrices generated from the rsfMRI graphs. Training of all models is done for 400 epochs using 5-fold cross-validation. Parameters is the number of parameters in all layers before the last FC.}
\label{mri}
\begin{tabular}{|c|c|c|}
\hline
Architecture & Accuracy(\%) & Parameters \\ \hline
FC           & $75.98 \pm2.05$ & $-$ \\ \hline
FC-FC        & $78.05 \pm2.55$ & 16384 \\ \hline \hline
32F-FC       & $81.09 \pm2.02$ & 64 \\ \hline
64F-FC       & $81.21 \pm3.59$ & 128 \\ \hline
2$\times$(32F)-FC   & $\textbf{83.78} \pm3.89$ & 128 \\ \hline
3$\times$(32F)-FC   & $82.68 \pm3.35$  & 192 \\ \hline
4$\times$(32F)-FC   & $81.82 \pm3.64$  & 256 \\ \hline
\end{tabular}
\end{table}



\subsection{Chemical Compound Classification}
\begin{figure}[!ht]
  \centering
  \includegraphics[width=0.8\figwidth]{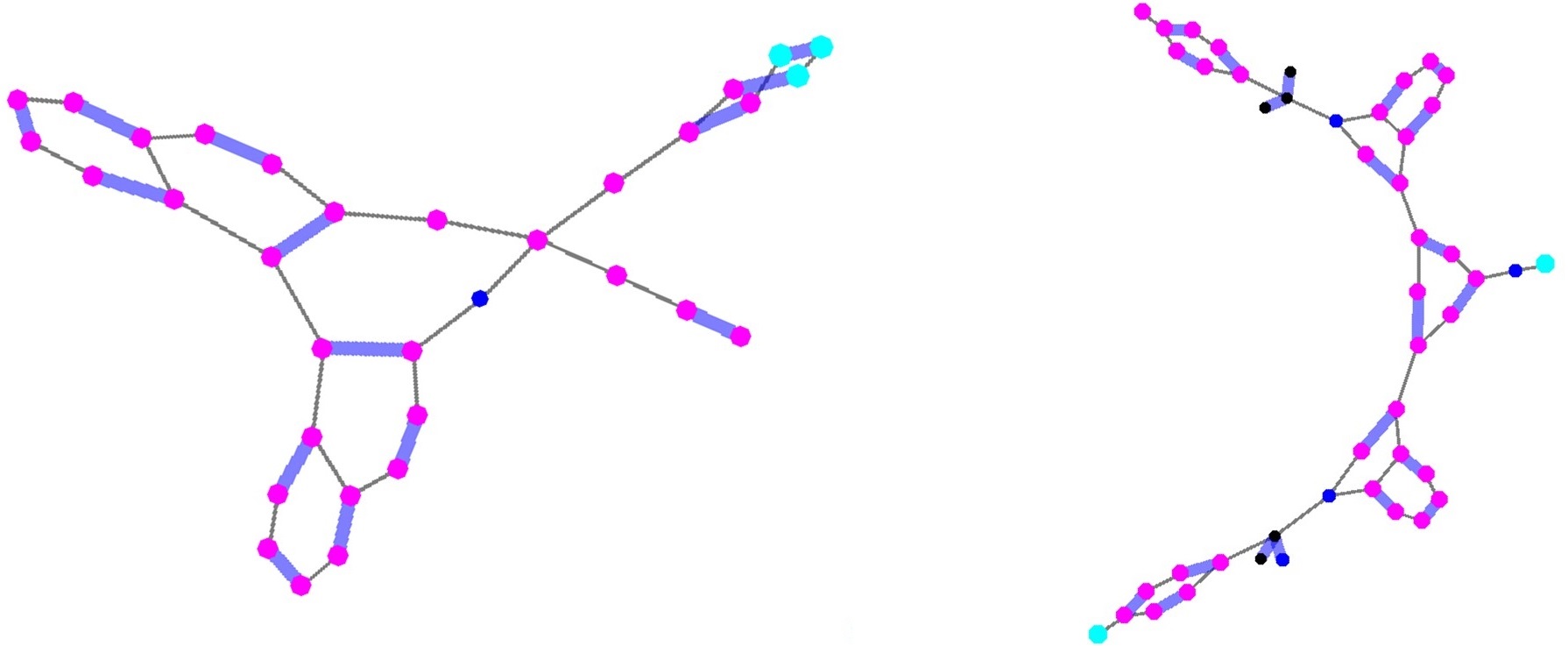}
  \caption{Samples of chemical compounds screened for activity against non-small cell lung cancer from the NCI1 dataset. Different colors and sizes represent different vertex features and edge features. \textbf{Left:} Negative sample. \textbf{Right:} Positive sample.}
\label{fig:protein1}
\end{figure}
A popular task in the pharmaceutical industry is classification and retrieval of chemical compounds.  The most common approach has been to use descriptors as inputs to a classifier. These descriptors can be individual fingerprints or substructures detected through a data mining step \cite{wale2008comparison}. Recently, there has been some work to build an end-to-end model capable of learning the best descriptors or a suitable classifier \cite{niepert2016cnn, atwood2015diffusion}. Deep learning based models were also proposed that extracts sub-structures by learning latent representations \cite{yanardag2015deep}. The main challenge is the inability to use a fully connected layer as a classifier since the size of input graphs is not constant. To address this, we used the graph embed pooling representation from Section \ref{subsec:graphpool}.

We use Graph-CNNs to address this problem and use the standard benchmark datasets $-$ NCI1 and D\&D, to compare classification performance.  NCI1 is a balanced graph dataset of chemical compounds that are screened for activity against non-small cell lung cancer and ovarian cancer cell lines respectively \cite{wale2008comparison}.  D\&D graphs are protein structures that can be classified into enzymes or non-enzymes categories \cite{dobson2003distinguishing}.  These data are highly complex in terms of size and structure of individual samples.  Each graph sample is heterogeneous and contains multiple adjacency matrices which indicate the presence of a specific bond type between two molecules. Detailed statistics and classification results on these datasets are listed in Table \ref{protein}. The Graph-CNN architectures achieve state-of-the-art performance compared with other recent approaches.

\begin{table}[!ht]
\centering
\caption{Comparison of classification accuracy for the chemical compound datasets. 5-hop DCNN $^{\#}$ accuracies are over 3-fold cross validation; all other accuracies are reported over 10-fold cross validation. GK $^*$ and WL $^*$ results as reported in \cite{niepert2016cnn}.}
\label{protein}
\begin{tabularx}{\columnwidth}{|X|c|c|}
\hline
Data set & NCI1 & D\&D \\ \hline
Maximum graph size & 111 & 5748 \\ \hline
Average graph size & 29.87 & 284.32  \\ \hline
\# Graphs & 4110 & 1178 \\ \hline
\hline
GK $^*$ \cite{shervashidze2009efficient} & 62.28 $\pm$ 0.29 & 78.45 $\pm$ 0.26  \\ \hline
WL $^*$ \cite{shervashidze2011weisfeiler} & 80.22 $\pm$ 0.51 & 77.95 $\pm$ 0.70  \\
\hline
PSCN \cite{niepert2016cnn} & $78.59 \pm 1.89$ & $77.12 \pm 2.41$ \\ \hline
Deep GK \cite{yanardag2015deep} & 80.31 $\pm$ 0.46 & $-$ \\ \hline
3$\times$16F-3x32F-GFC32 & $83.69 \pm 1.40$ & $-$ \\ \hline
6$\times$32F-GFC32 & $83.57 \pm 1.99$ & $-$ \\ \hline
2$\times$64F-Pool32-FC256 & $84.08 \pm 1.45$ & $-$ \\ \hline
2$\times$64F-Pool32-32F-Pool8-FC256 & $\textbf{84.62} \pm 2.24$ & $81.45\pm 2.87$ \\ \hline
2$\times$64F-Pool32-32F-Pool8-64F-FC256 & $83.48 \pm 1.36$ & $-$ \\ \hline
2$\times$64F-Pool32-64F-Pool8-FC256 & $84.35 \pm 1.00$ & $\textbf{81.88} \pm 3.39$ \\ \hline
\hline
5-hop DCNN $^{\#}$ \cite{atwood2015diffusion} & 62.61 & $-$ \\ \hline
2$\times$64F-Pool32-32F-Pool8-FC256 $^{\#}$ & $\textbf{81.98} \pm 0.76$ & $-$ \\ \hline
\end{tabularx}
\end{table}

\subsection{Bosphorus 3D Facial Expressions}

One common form of heterogeneous graph data is 3D mesh data. While sensors generally collect 3D data as point clouds \cite{bosphorus3d} or images with a depth channel \cite{largeobjectset}, these types of data can often be used to construct 3D meshes, posed as vertices and edges \cite{meshlab,largeobjectset}. Graph-CNN can perform object recognition tasks on 3D mesh data. This could be applied to autonomous driving applications that depend on LiDAR point clouds.

\begin{figure}[!ht]
  \centering
  \includegraphics[width=\figwidth]{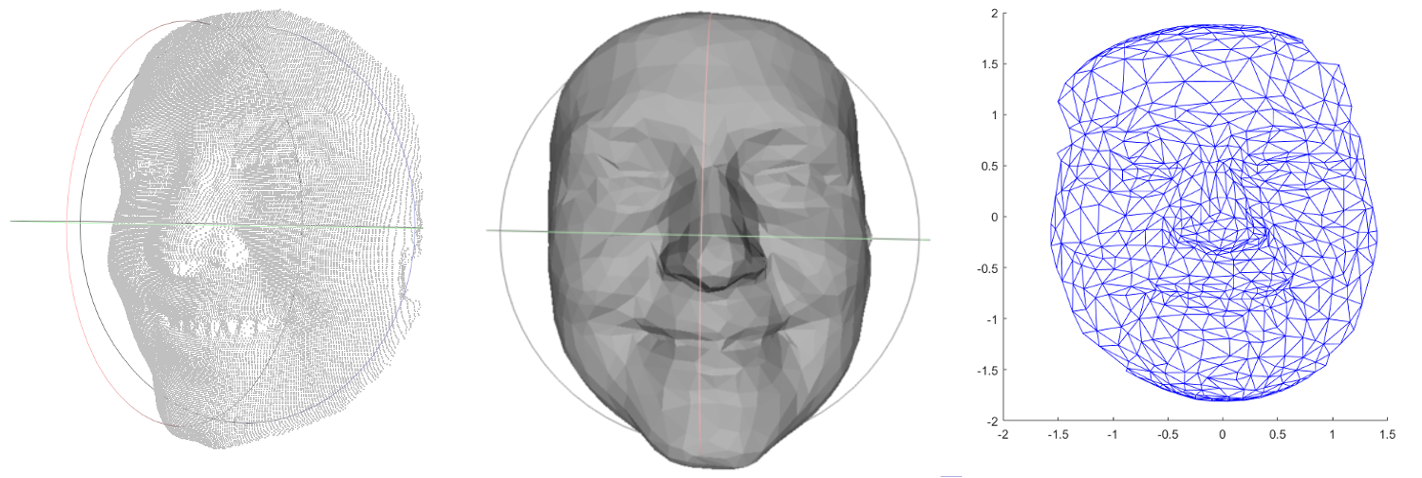}
  \caption{Different stages of Bosphorus face preprocessing. \textbf{Left:} Original point cloud. \textbf{Center:} Low-poly 3D mesh. \textbf{Right:} 2D projection of the front of 3D mesh that we input into Graph CNN.}
\label{fig:bosphigure}
\end{figure}

We test Graph-CNN on a toy 3D mesh classification problem. We use the facial expression-labeled faces in the Bosphorus 3D Face dataset \cite{bosphorus3d}. 453 of the provided 3D faces are labeled with one of six facial expressions: Anger, Disgust, Fear, Happiness, Sadness, and Surprise.  Bosphorus data is encoded as point cloud data, but with tools such as Meshlab \cite{meshlab} can be converted to 3D meshes. We attempt to classify these faces with Graph-CNN. RGB-D datasets were considered \cite{largeobjectset,NYUv2,B3DO,RGBD}, but RGB-D data is much less trivial to convert to a 3D mesh than point clouds. In addition, we were ultimately seeking a dataset that had focused samples neatly divided into a set of classes, similar to ImageNet \cite{russakovsky2015imagenet}.

It may seem curious to seasoned computer graphics experts that we formulated these 3D shapes as graphs, using only vertex and edge data. Other works treat 3D shapes as continuous manifolds, or discretely as a tuple $(\boldsymbol{V},\boldsymbol{E},\boldsymbol{F})$, where $\boldsymbol{E}$ is the set of edges and $\boldsymbol{F}$ is the set of faces \cite{masci2015geodesic,boscaini2016classdesc,boscaini2016anisotropicdesc,boscaini2016anisotropiccnn,monti2016geometric,bronstein2016geometricdeep}. The purpose of this experiment was less to compete with the state-of-the-art in 3D shape retrieval and more to show the versatility of Graph-CNNs by applying them to a nontraditional graph dataset.


We discuss the preprocessing for this problem in the appendix in \ref{subsec:bosphoruspreproc}. We first evaluate our dataset by creating multiple Graph-CNN networks with different convolutional layers. Each Graph-CNN layer (labeled GConv in Table \ref{tab:bosgconvall}) learns 16 filters. Each network ends in a graph embed pooling layer as described in Section \ref{subsec:graphpool}, resulting in a new 32-vertex graph that is input into a fully-connected layer. Pooling was not used to reduce the data between graph convolutions. Table \ref{tab:bosgconvall} shows the results. The mean accuracy after 5-fold cross validation is displayed, plus or minus one standard deviation from the mean. Three Graph-CNN layers appears to underfit, and five appears to overfit the data. Since pooling only occurs at the end, and only as a method to fit heterogeneous data to a fixed fully-connected layer, the potential depth of this network is limited.

\begin{table}[!ht]
\centering
\caption{Effect of Network Depth on Accuracy.}
\label{tab:bosgconvall}
\begin{tabular}{|c|c|c|}
\hline
Architecture & Accuracy      & Parameters \\ \hline
3$\times$ GConv16    & $54.8\pm4.51\%$ & 8016       \\ \hline
4$\times$ GConv16    & $70.0\pm9.04\%$ & 10336      \\ \hline
5$\times$ GConv16    & $67.9\pm7.33\%$ & 12656      \\ \hline
\end{tabular}
\end{table}

We also evaluate three other methods for comparison, also with 5-fold cross validation.  First, we transform the 2D vertex data into PCA space and classify it with a Support Vector Machine (SVM). Second, we transform the 2D vertex data into Locality-Preserving Projection space (LPP) \cite{niyogi2004locality}, and pass that through an SVM. Finally we create a more typical 5 convolution layer CNN architecture (with slightly fewer total parameters) used for processing images. Each 3D face in Bosphorus has a corresponding image of that face, so we train the CNN on those images. Each layer in the CNN architecture is a $3\times 3$ convolution, followed by a max-pooling with a stride of 2. The first 3 convolution layers have 16 filters, the last two have 8. At the end of these layers is a fully-connected layer with 6 outputs, and a softmax. Batch normalization, ReLUs, and the same training hyperparameters as Graph-CNN are used.

The final results of all these experiments are in Table \ref{tab:bosresults}. The SVM classifiers do not model the geometric data well. The traditional CNN with image data beats the Graph-CNN data, but its data does have some richer features. The input into the Graph-CNN does not encode color or depth. In addition, our network has some limitations. Our receptive field may be limited due to the shallowness of the network and the lack of pooling. Efforts to increase this field, such as pooling, increased depth, or atrous filtering techniques as in \cite{deeplab} would likely yield superior results. Also, no data augmentation was attempted. Augmenting the dataset with affine transforms of the vertices could address the small size of the dataset. Regardless, we have shown that Graph-CNNs are capable of processing 3D mesh data, and have plenty of room for iteration to become competitive.  

\begin{table}[!ht]
\centering
\caption{Bosphorus Expression Results.}
\label{tab:bosresults}
\begin{tabular}{|c|c|c|}
\hline
Method     & Accuracy (\%)   & Parameters \\ \hline
PCA+SVM    & 28.0            & 283        \\ \hline
LPP+SVM    & 26.2            & 5          \\ \hline
CNN+Images & \textbf{82.2} & 8032       \\ \hline
Graph-CNN  & 70.0            & 10336      \\ \hline
\end{tabular}
\end{table}

\subsection{Document Classification}
A common form of graph-formatted data is a network of documents.
For example, scientific documents in a database are related to one another through citations and references. The entire network is a single large graph, rather than a set of disparate graphs. Each document is a vertex in the graph with a certain features and a citation is an edge from one vertex to the other. Administrators of such large networks may desire to automatically label documents according to their relationship to the rest of the literature.
We demonstrate the use of Graph-CNN architecture to model such a vertex classification task. Since the data is organized as a single graph, a label mask of zeros is applied on the test vertices during training. Hence, the loss layer does not back-propagate for the test vertices. At test time, only the test labels are used to compute the accuracy.


Evaluation of Graph-CNN for such an application is focused on the Cora dataset \cite{docclassdata}, a large network of scientific publications connected through citations. The vertex features in this case are binary word vectors that indicate the presence of a word from a dictionary of 1433 unique words. There are 2708 publications classified under seven different categories- Case Based, Genetic Algorithms, Neural Networks, Probabilistic Methods, Reinforcement Learning, Rule Learning and Theory. There is an edge connection from a cited article to a citing article and another edge connection from a citing article to a cited article. These edge features are binary representations.  We perform cross validations with three different settings to form the training and test set for fair comparison with other recent studies.

Table \ref{tab:cora_comp} lists document classification accuracies compared with the recent approaches. Our Graph-CNN architecture (2$\times$48F-7F) contains three Graph-CNN layers: first two layers with 48 filters and third layer with seven filters. The last Graph-CNN layer computes the prediction of each vertex. We then expand this network by adding 0-hop filters after each Graph-CNN. Dropout was also added before each 0-hop filter. We note our performance on the 1000 test* is not state-of-the-art.  This test split has only 20 training samples per class and our method overfits, achieving 100\% classification accuracy on the training set after only three iterations.  We observed that with deeper architectures (3x48F-7F), the network quickly overfits on the training set and the performance degrades on the test set. We report these in Table \ref{tab:docparams}. For the model with Dropout and 0$-$hop, the highest accuracy that we obtain are 89.14\% and 91.51\% on 3- and 10- folds, respectively. All models were trained using Adam optimization \cite{kingma2014adam} and identical hyperparameters. The BatchNorm layers were modified to no longer use running average for mean and variance since there is only a single large sample graph.  

\begin{table}[!ht]
\centering
\caption{Cora document classification accuracy.}
\label{tab:docparams}
\begin{tabular}{|c|c|c|}
\hline
Method                                  & \multicolumn{2}{c|}{Accuracy (\%)}                        \\ \hline
                       & 3-fold                     & 10-fold \\ \hline
2$\times$48F-7F                                & 84.30 $\pm$ 1.66            & 87.11 $\pm$ 1.84  \\ \hline
+ Dropout, 0$-$hop                      & \textbf{87.55} $\pm$ 1.38 & \textbf{89.18} $\pm$ 1.96   \\ \hline
3x48F-7F & 84.86 $\pm$ 0.42 & 87.44 $\pm$ 1.83   \\ \hline
\end{tabular}
\end{table}

\begin{table}[!ht]
\centering
\caption{Cora document classification accuracy comparison. 3-fold and 10-fold are cross validation tests, and \enquote{1000 test} is a popular split in the literature where 1000 samples are used in the test set, and the rest are used in training. 
\enquote{1000 test*} refers to an experiment split using only 20 samples from each class for training, and the same 1000 samples for testing.}
\label{tab:cora_comp}
\begin{tabular}{|c|c|c|}
\hline
Method                            & Split     & Accuracy \\ \hline
Yang \cite{yang2016revisiting}                & 1000 test & 75.7     \\ \hline
Kipf \cite{kipf2016semisupervised}                      & 1000 test* & 81.5     \\ \hline
Monti \cite{monti2016geometric}                & 1000 test* & \textbf{81.69}    \\ \hline
DCNN \cite{atwood2015diffusion} & 3-fold    & 86.77    \\ \hline
Ours                              & 1000 test*   & 76.3  \\ \hline
Ours                              & 1000 test & \textbf{86.56} $\pm$ 0.68    \\ \hline
Ours                              & 3 fold    & \textbf{87.55} $\pm$ 1.38   \\ \hline
Ours                              & 10 fold   & \textbf{89.18} $\pm$ 1.96   \\ \hline
\end{tabular}
\end{table}

\section{Computational Complexity}
For a system to be a feasible solution to deep learning applications it must perform well not only in terms of accuracy but also in terms of required computational resources. It must also be efficiently computed using general purpose computation resources like CPUs and GPUs. The worst case scenario of our system is an application with fully-connected graph samples, where each sample has $N$ vertices, $L$ adjacencies, $C$ vertex features, and $F$ filters. The time and space complexity for (\ref{equ:onetensorfilter}) is $O(N^2CLF)$ and $O(N^2CF + N^2L)$ respectively. It is trivial to see that the neighborhood of a node can be computed prior to applying specific filter weights. This modification is seen in (\ref{equ:neighbor_calc}) and (\ref{equ:v_neighbor_calc}), where $\mathcal{N}^{(\ell)}$ is computed once per sample with $\mathcal{N}$ in $\mathbb{R}^{N\times CL}$ and $h$ (the weight matrix) in $\mathbb{R}^{CL\times F}$. Equation (\ref{equ:neighbor_calc}) is analogous to an \textit{im2col} operation commonly used in CNN computations where input features are organized before a filter is applied. In fact, for images, the result is the same as an \textit{im2col} operation. If neighborhood is computed before applying the filter, the time and space complexity becomes $O(N^2CL + NCLF)$ and $O(N^2L + NCL + NF)$ respectively. Note that (\ref{equ:onetensorfilter}) has the benefit of having a reusable $\boldsymbol{H}$ for all samples with the same adjacency matrices (e.g. images) reducing the complexity of batched operations on homogeneous problems.
    
\begin{equation}
\label{equ:neighbor_calc}
\mathcal{N}^{(\ell)} = \boldsymbol{A}_\ell\boldsymbol{V}_{in}
\end{equation}

\begin{equation}
\label{equ:v_neighbor_calc}
\boldsymbol{V}_{out} = \mathcal{N}h
\end{equation}
    
    When using sparsely connected graphs, the adjacency matrices can be sparsely represented. Doing so reduces time and space complexity to $O(EC + NCLF)$ and $O(E + NCL + NF)$ respectively where $E$ is the number of non-zero edge features ($E < N^2L$). For a CNN, each vertex has at most one neighbor in each adjacency matrix so $E \approx LN$ and the time complexity of Graph-CNN is reduced to $O(NCLF)$ which is equivalent to that of a CNN.

\section{Conclusion}
Many types of graph data are heterogeneous and cannot be processed using traditional spectral convolutional filtering techniques.  We introduce a general purpose Graph-CNN paradigm that offers the same breakthrough benefits currently only afforded to homogeneous data. Similar to traditional CNN architectures, Graph-CNNs operate directly in the spatial domain to generate semantically rich features.  Our model operates on both homogeneous and heterogeneous data, learning properties from both graph vertices and edges. We establish that traditional CNNs are a subset of Graph-CNN for image data. We have proposed a graph embed pooling method that can reduce dimensionality of graphs throughout a network. We have shown results on graphs of fixed size and connections (images), graphs with fixed size but variable connections (rsfMRI), graphs with varying size and connections (chemical compounds, facial expression recognition), and large single-sample graphs (document classification).

Future work involves extending the flexibility and applications of our Graph-CNN method. We seek to increase the depth and receptive field of our networks through more sophisticated pooling methods, residual network formulations, and atrous filtering.  The mechanics of these Graph-CNNs should be analyzed through filter visualization and more in-depth study of the distributions of graph data across the network. The theory should be expanded to enable filtering of edges as well as vertices.  We anticipate methods such as Graph-CNN will be applied far and wide to bring the benefits of automatic feature learning to graph problems throughout the literature. 

\section{Appendix}

\subsection{Graph-CNN is a Superset of CNNs}
\label{subsec:proof}
\def\matmult{}
\def\compmult{\odot}
\def\widthsign{W}

\def\checkmark{\tikz\fill[scale=0.4](0,.35) -- (.25,0) -- (1,.7) -- (.25,.15) -- cycle;}
In this section we prove that Graph-CNN is a superset of CNNs.


A $3\times 3$ convolutional filtering operation on a single $3\times 3$ neighborhood can be written as in (\ref{eqn:imagefir}).
\begin{equation}
\begin{multlined}
\label{eqn:imagefir}
V'_{i,j} = h_0\cdot V_{i,j}\\+h_1\cdot V_{i-1,j-1}+h_2\cdot V_{i-1,j}\\+h_3\cdot V_{i-1,j+1}+h_4\cdot V_{i,j-1} \\ + h_5\cdot V_{i,j+1}+h_6\cdot V_{i+1,j-1}\\+h_7\cdot V_{i+1,j}+h_8\cdot V_{i+1,j+1}
\end{multlined}
\end{equation}

Vertex features in Graph-CNN are represented as a one-dimensional vector. To pose two-dimensional pixels as a vertex feature vector, we can use the following mapping from $(i,j) \rightarrow n$.

\begin{equation}
n = i \cdot W + j
\end{equation}

\begin{table}[!ht]
\centering
\begin{tabular}{|c|c|c|}
\hline
1 & 2 & 3 \\
\hline
4 & 5 & 6 \\
\hline
7 & 8 & 9 \\
\hline
\end{tabular}
\end{table}

\begin{table}[!ht]
\centering
\begin{tabular}{|c|c|c|c|c|c|c|c|c|}
\hline
1 & 2 & 3 &4 & 5 & 6 & 7 & 8 & 9 \\
\hline
\end{tabular}
\end{table}

Where W is the horizontal width of the image. Equation (\ref{eqn:imagefir}) can be reposed in this indexing scheme as (\ref{eqn:imagefir1dindex}).

\begin{equation}
\begin{multlined}
\label{eqn:imagefir1dindex}
V'_n = h_0\cdot V_n\\+h_1\cdot V_{n-\widthsign-1}+h_2\cdot V_{n-\widthsign} \\+h_3\cdot V_{n-\widthsign+1}+h_4\cdot V_{n-1} \\+  h_5\cdot V_{n+1}+h_6\cdot V_{n+\widthsign-1}\\+h_7\cdot V_{n+\widthsign}+h_8\cdot V_{n+\widthsign+1} 
\end{multlined}
\end{equation}

Next, we define a series of adjacency matrices, each one only containing the pixel-wise relationships of a certain direction, as illustrated in the right-side image in Figure \ref{fig:filter}.
\begin{equation}
\uparrow \boldsymbol{A}_{i,j} = 
\begin{cases}
    1,& \text{if } i-\widthsign = j \\
    0,              & \text{otherwise}
\end{cases}
\end{equation}
\begin{equation}
\downarrow \boldsymbol{A}_{i,j} = \uparrow \boldsymbol{A}_{j,i} = \uparrow \boldsymbol{A}_{i,j}^T =
\begin{cases}
    1,& \text{if } i + \widthsign = j \\
    0,              & \text{otherwise}
\end{cases}
\end{equation}

\begin{equation}
\leftarrow \boldsymbol{A}_{i,j} = 
\begin{cases}
    1,& \text{if } i-1 = j \\
    0,              & \text{otherwise}
\end{cases}
\end{equation}
\begin{equation}
\rightarrow \boldsymbol{A}_{i,j} = \leftarrow \boldsymbol{A}_{j,i} =
\begin{cases}
    1,& \text{if } i + 1= j \\
    0,              & \text{otherwise}
\end{cases}
\end{equation}

\begin{equation}
\nwarrow \boldsymbol{A}_{i,j} = 
\begin{cases}
    1,& \text{if } i-\widthsign-1 = j \\
    0,              & \text{otherwise}
\end{cases}
\end{equation}
\begin{equation}
\searrow \boldsymbol{A}_{i,j} = \nwarrow \boldsymbol{A}_{j,i} =
\begin{cases}
    1,& \text{if } i+\widthsign + 1= j \\
    0,              & \text{otherwise}
\end{cases}
\end{equation}

\begin{equation}
\nearrow \boldsymbol{A}_{i,j} = 
\begin{cases}
    1,& \text{if } i-\widthsign+1 = j \\
    0,              & \text{otherwise}
\end{cases}
\end{equation}
\begin{equation}
\swarrow \boldsymbol{A}_{i,j} = \nearrow \boldsymbol{A}_{j,i} =
\begin{cases}
    1,& \text{if } i+\widthsign - 1= j \\
    0,              & \text{otherwise}
\end{cases}
\end{equation}

\begin{equation}
\boldsymbol{I} =
\begin{cases}
    1,& \text{if } i = j \\
    0,              & \text{otherwise}
\end{cases}
\end{equation}

These matrices are now indicator functions to clarify which filter taps apply to which pixels in a given pixel neighborhood. This means that we can remove the indices entirely from (\ref{eqn:imagefir1dindex}) and pose it as a matrix operation, as in (\ref{equ:cnn}).

\begin{equation}
\label{equ:cnn}
\begin{multlined}
\boldsymbol{V}' = h_0\cdot \colj{\boldsymbol{I}} \boldsymbol{V}\matmult \\+h_1\cdot\colj{\nwarrow \boldsymbol{A}} \boldsymbol{V}\matmult +h_2\cdot \colj{\uparrow \boldsymbol{A}}\boldsymbol{V}\matmult \\+h_3\cdot \colj{\nearrow \boldsymbol{A}}\boldsymbol{V}\matmult +h_4\cdot\colj{\leftarrow \boldsymbol{A}}\boldsymbol{V}\matmult  \\ + h_5\cdot \colj{\rightarrow \boldsymbol{A}} \boldsymbol{V}\matmult+h_6\cdot\colj{\swarrow \boldsymbol{A}} \boldsymbol{V}\matmult \\+h_7\cdot \colj{\downarrow \boldsymbol{A}} \boldsymbol{V}\matmult+h_8\cdot \colj{\searrow \boldsymbol{A}}\boldsymbol{V}\matmult  \\
\end{multlined}
\end{equation}

The separate adjacency matrices can be packed into a tensor $\boldsymbol{\mathcal{A}}$ made up of the slices $\{ \boldsymbol{I}, \nwarrow \boldsymbol{A}, \uparrow \boldsymbol{A}, \nearrow \boldsymbol{A}, \leftarrow \boldsymbol{A}, \rightarrow \boldsymbol{A}, \swarrow \boldsymbol{A}, \downarrow \boldsymbol{A}, \searrow \boldsymbol{A}\}$. At this point it can be processed by equations (\ref{equ:linearfiltermultipletensormultiplefeature}) and (\ref{equ:onetensorfilter}).

\subsection{Spectral Domain}
\label{subsection:SpectralDomain}
In \cite{henaff2015deep}, experiments on ImageNet classification are performed to compare learning for spectral graph convolutions and traditional image convolutions. Their results show that spectral graph convolutions learn more quickly than image convolutions at first, but ultimately converge to the same result.

\begin{figure}[!ht]
  \centering
  \includegraphics[width=0.8\figwidth]{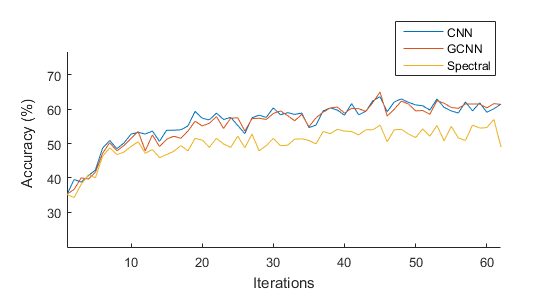}
  \caption{Comparison of accuracy for CNN, Graph-CNN, and Graph-Spectral methods trained on CIFAR-10 dataset. All architectures were single layer convolution with 32 filters followed by two fully connected layers.}
  \label{fig:spectralcompare}
\end{figure}

In this section, we compare image convolutions (CNN), spectral convolutions (Spectral), and our Graph-CNN (GCNN). We trained a single-layer convolutional neural network of each of the three types on the CIFAR-10 dataset \cite{krizhevsky2009learning}. Each network learned 32 filters, and was followed by two fully-connected layers of equal size.  The resulting learning curve is shown as in Figure \ref{fig:spectralcompare}. 
Since we proved in \ref{subsec:proof} that an image CNN can be exactly represented by Graph-CNN, these two methods understandably get the same results. However, our spectral performance in this test case is about 10\% lower than the other methods. These findings contrast with the results of \cite{henaff2015deep} which get identical performance between spectral and image convolutions. Several hyperparameters were explored, but it is possible further hyperparameter tuning could improve the Graph-Spectral performance.
\subsection{Preparing the Bosphorus Model}
\label{subsec:bosphoruspreproc}
One of the biggest challenges with this dataset as a testbed for Graph-CNN is its small number of samples, each with a very high dimensionality (tens of thousands of points in 3D space). This means that the data, without upfront reduction, can rapidly lead to overfitting and poor generalization in a network. In addition, adjacency matrices built from this many points would be expensive to compute.  For this experiment we performed extensive data reduction, attempting to cut away as much detail from these point clouds to enable effective and efficient learning.

First, each face is processed as a raw point cloud. Outlier points are removed and a mesh is created from the remainder using Meshlab \cite{meshlab}. The mesh is simplified to a low dimension (roughly $1052$ vertices) using Meshlab's mesh simplification algorithms. The mesh is then converted into a $V$ and $A$ matrix. The features of $V$ are the X, Y, and Z coordinates in 3D Euclidean space of each vertex.  The edges of $A$ are $1$ if the edge exists in the mesh and $0$ if it does not.  We did not use a distance measure such as Euclidean distance because that information would be implicit in the vertex features. To increase the number of edge features, we partitioned the edges into bins based on the angle of the vector formed by that edge. Edge $ij$ forms a vector $\vec{ij}$ defined by subtracting $i$ from $j$. Then the sign of each component ($ij_X$,$ij_Y$, and $ij_Z$) from that vector is used to separate into bins: ($ij_X \geq 0$,$ij_Y \geq 0$, $ij_Z \geq 0$) is one bin, ($ij_X \geq 0$,$ij_Y \geq 0$, $ij_Z < 0$) is another bin, and so on for all eight 3D octants.

Next we attempt to further reduce dimensionality by removing the back of the head formed by the mesh creation algorithm. These vertices and edges are generated in an attempt to create a closed 3D figure, but they provide no information for discerning facial expression. To do this programmatically, we normalize $V$ for a given sample to be $0$ mean, unit variance. Vertices with negative Z components are removed from $V$ and $A$. Finally, we project the face onto a 2D-plane by removing the Z feature from $V$. The 8 adjacency matrices are still preserved, though the $Z$ feature is removed. This much reduced model of the face retains useful information for identifying expression. Figure \ref{fig:bosphigure} illustrates the transformation.
\section{Acknowledgements}
We would like to thank the National GEM Consortium for funding one of the student authors through the GEM Fellowship.
\ifCLASSOPTIONcaptionsoff
  \newpage
\fi



\bibliographystyle{IEEEtran}
%
\bibliography{egbib}

\begin{thebibliography}{10}
\providecommand{\url}[1]{#1}
\csname url@samestyle\endcsname
\providecommand{\newblock}{\relax}
\providecommand{\bibinfo}[2]{#2}
\providecommand{\BIBentrySTDinterwordspacing}{\spaceskip=0pt\relax}
\providecommand{\BIBentryALTinterwordstretchfactor}{4}
\providecommand{\BIBentryALTinterwordspacing}{\spaceskip=\fontdimen2\font plus
\BIBentryALTinterwordstretchfactor\fontdimen3\font minus
  \fontdimen4\font\relax}
\providecommand{\BIBforeignlanguage}[2]{{%
\expandafter\ifx\csname l@#1\endcsname\relax
\typeout{** WARNING: IEEEtran.bst: No hyphenation pattern has been}%
\typeout{** loaded for the language `#1'. Using the pattern for}%
\typeout{** the default language instead.}%
\else
\language=\csname l@#1\endcsname
\fi
#2}}
\providecommand{\BIBdecl}{\relax}
\BIBdecl

\bibitem{krizhevsky2012imagenet}
A.~Krizhevsky, I.~Sutskever, and G.~E. Hinton, ``Imagenet classification with
  deep convolutional neural networks,'' in \emph{Advances in neural information
  processing systems}, 2012, pp. 1097--1105.

\bibitem{simonyan2014very}
K.~Simonyan and A.~Zisserman, ``Very deep convolutional networks for
  large-scale image recognition,'' \emph{arXiv preprint arXiv:1409.1556}, 2014.

\bibitem{Szegedy_2015_CVPR}
C.~Szegedy, W.~Liu, Y.~Jia, P.~Sermanet, S.~Reed, D.~Anguelov, D.~Erhan,
  V.~Vanhoucke, and A.~Rabinovich, ``Going deeper with convolutions,'' in
  \emph{The IEEE Conference on Computer Vision and Pattern Recognition (CVPR)},
  June 2015.

\bibitem{He2015}
K.~He, X.~Zhang, S.~Ren, and J.~Sun, ``Deep residual learning for image
  recognition,'' \emph{arXiv preprint arXiv:1512.03385}, 2015.

\bibitem{renNIPS15fasterrcnn}
S.~Ren, K.~He, R.~Girshick, and J.~Sun, ``Faster {R-CNN}: Towards real-time
  object detection with region proposal networks,'' in \emph{Advances in Neural
  Information Processing Systems ({NIPS})}, 2015.

\bibitem{redmon2015you}
J.~Redmon, S.~Divvala, R.~Girshick, and A.~Farhadi, ``You only look once:
  Unified, real-time object detection,'' \emph{arXiv preprint
  arXiv:1506.02640}, 2015.

\bibitem{abdelhamid2014asr}
\BIBentryALTinterwordspacing
O.~Abdel-Hamid, A.-R. Mohamed, H.~Jiang, L.~Deng, G.~Penn, and D.~Yu,
  ``Convolutional neural networks for speech recognition,'' \emph{IEEE/ACM
  Trans. Audio, Speech and Lang. Proc.}, vol.~22, no.~10, pp. 1533--1545, Oct.
  2014. [Online]. Available: \url{http://dx.doi.org/10.1109/TASLP.2014.2339736}
\BIBentrySTDinterwordspacing

\bibitem{henaff2015deep}
M.~Henaff, J.~Bruna, and Y.~LeCun, ``Deep convolutional networks on
  graph-structured data,'' \emph{arXiv preprint arXiv:1506.05163}, 2015.

\bibitem{bruna2013spectral}
J.~Bruna, W.~Zaremba, A.~Szlam, and Y.~LeCun, ``Spectral networks and locally
  connected networks on graphs,'' \emph{arXiv preprint arXiv:1312.6203}, 2013.

\bibitem{kipf2016semisupervised}
\BIBentryALTinterwordspacing
T.~N. Kipf and M.~Welling, ``Semi-supervised classification with graph
  convolutional networks,'' in \emph{ArXiv e-prints}, vol. 1609, Conference
  Proceedings. [Online]. Available:
  \url{http://adsabs.harvard.edu/abs/2016arXiv160902907K}
\BIBentrySTDinterwordspacing

\bibitem{ktena2017distance}
S.~I. Ktena, S.~Parisot, E.~Ferrante, M.~Rajchl, M.~Lee, B.~Glocker, and
  D.~Rueckert, ``Distance metric learning using graph convolutional networks:
  Application to functional brain networks,'' \emph{arXiv preprint
  arXiv:1703.02161}, 2017.

\bibitem{Defferrard2016}
M.~Defferrard, X.~Bresson, and P.~Vandergheynst, ``Convolutional neural
  networks on graphs with fast localized spectral filtering,'' in
  \emph{Advances in Neural Information Processing Systems 29}, D.~D. Lee,
  M.~Sugiyama, U.~V. Luxburg, I.~Guyon, and R.~Garnett, Eds.\hskip 1em plus
  0.5em minus 0.4em\relax Curran Associates, Inc., Conference Proceedings, pp.
  3844--3852.

\bibitem{sandryhaila2013discrete}
A.~Sandryhaila and J.~M. Moura, ``Discrete signal processing on graphs,''
  \emph{Signal Processing, IEEE Transactions on}, vol.~61, no.~7, pp.
  1644--1656, 2013.

\bibitem{Felipe2017}
F.~Petroski~Such, ``Graphcnn,'' \url{https://github.com/fps7806/Graph-CNN},
  2017.

\bibitem{edwards2016graphcnn}
\BIBentryALTinterwordspacing
M.~Edwards and X.~Xie, ``Graph convolutional neural network,'' in \emph{British
  Machine Vision Conference}, R.~C. Wilson, E.~R. Hancock, and W.~A.~P. Smith,
  Eds.\hskip 1em plus 0.5em minus 0.4em\relax BMVA Press, Conference
  Proceedings. [Online]. Available:
  \url{http://www.bmva.org/bmvc/2016/papers/paper114/paper114.pdf}
\BIBentrySTDinterwordspacing

\bibitem{vonLuxburg2007}
\BIBentryALTinterwordspacing
U.~von Luxburg, ``A tutorial on spectral clustering,'' \emph{Statistics and
  Computing}, vol.~17, no.~4, pp. 395--416, 2007. [Online]. Available:
  \url{http://dx.doi.org/10.1007/s11222-007-9033-z}
\BIBentrySTDinterwordspacing

\bibitem{dhillon2007}
I.~S. Dhillon, Y.~Guan, and B.~Kulis, ``Weighted graph cuts without
  eigenvectors a multilevel approach,'' \emph{IEEE Transactions on Pattern
  Analysis and Machine Intelligence}, vol.~29, no.~11, pp. 1944--1957, 2007.

\bibitem{Shuman2016}
\BIBentryALTinterwordspacing
D.~I. Shuman, B.~Ricaud, and P.~Vandergheynst, ``Vertex-frequency analysis on
  graphs,'' \emph{Applied and Computational Harmonic Analysis}, vol.~40, no.~2,
  pp. 260--291, 2016. [Online]. Available:
  \url{http://www.sciencedirect.com/science/article/pii/S1063520315000214}
\BIBentrySTDinterwordspacing

\bibitem{seo2016graphrecurrent}
\BIBentryALTinterwordspacing
Y.~Seo, M.~Defferrard, P.~Vandergheynst, and X.~Bresson, ``Structured sequence
  modeling with graph convolutional recurrent networks,'' in \emph{ArXiv
  e-prints}, vol. 1612, Conference Proceedings. [Online]. Available:
  \url{http://adsabs.harvard.edu/abs/2016arXiv161207659S}
\BIBentrySTDinterwordspacing

\bibitem{lecun98}
Y.~Lecun, L.~Bottou, Y.~Bengio, and P.~Haffner, ``Gradient-based learning
  applied to document recognition,'' \emph{Proceedings of the IEEE}, vol.~86,
  no.~11, pp. 2278--2324, 1998.

\bibitem{atwood2015diffusion}
J.~Atwood and D.~Towsley, ``Diffusion-convolutional neural networks,''
  \emph{arXiv preprint arXiv:1511.02136}, 2015.

\bibitem{niepert2016cnn}
M.~Niepert, M.~Ahmed, and K.~Kutzkov, ``Learning convolutional neural networks
  for graphs,'' in \emph{Proceeding of the 33rd International Conference on
  Machine Learning}, 2016, pp. 2014--–2023.

\bibitem{DBLP:journals/corr/DuvenaudMAGHAA15}
\BIBentryALTinterwordspacing
D.~K. Duvenaud, D.~Maclaurin, J.~Aguilera{-}Iparraguirre,
  R.~G{\'{o}}mez{-}Bombarelli, T.~Hirzel, A.~Aspuru{-}Guzik, and R.~P. Adams,
  ``Convolutional networks on graphs for learning molecular fingerprints,''
  \emph{CoRR}, vol. abs/1509.09292, 2015. [Online]. Available:
  \url{http://arxiv.org/abs/1509.09292}
\BIBentrySTDinterwordspacing

\bibitem{ggnn2016}
\BIBentryALTinterwordspacing
Y.~Li, D.~Tarlow, M.~Brockschmidt, and R.~S. Zemel, ``Gated graph sequence
  neural networks,'' \emph{CoRR}, vol. abs/1511.05493, 2015. [Online].
  Available: \url{http://arxiv.org/abs/1511.05493}
\BIBentrySTDinterwordspacing

\bibitem{scarselli2009graph}
F.~Scarselli, M.~Gori, A.~C. Tsoi, M.~Hagenbuchner, and G.~Monfardini, ``The
  graph neural network model,'' \emph{Neural Networks, IEEE Transactions on},
  vol.~20, no.~1, pp. 61--80, 2009.

\bibitem{perozzi2014deepwalk}
B.~Perozzi, R.~Al-Rfou, and S.~Skiena, ``Deepwalk: Online learning of social
  representations,'' in \emph{Proceedings of the 20th ACM SIGKDD}.\hskip 1em
  plus 0.5em minus 0.4em\relax ACM, 2014, pp. 701--710.

\bibitem{masci2015geodesic}
J.~Masci, D.~Boscaini, M.~M. Bronstein, and P.~Vandergheynst, ``Geodesic
  convolutional neural networks on riemannian manifolds,'' in \emph{2015 IEEE
  International Conference on Computer Vision Workshop (ICCVW)}, Conference
  Proceedings, pp. 832--840.

\bibitem{boscaini2016classdesc}
D.~Boscaini, J.~Masci, S.~Melzi, M.~M. Bronstein, U.~Castellani, and
  P.~Vandergheynst, ``Learning class-specific descriptors for deformable shapes
  using localized spectral convolutional networks,'' in \emph{Proceedings of
  the Eurographics Symposium on Geometry Processing}.\hskip 1em plus 0.5em
  minus 0.4em\relax 2853911: Eurographics Association, Conference Proceedings,
  pp. 13--23.

\bibitem{boscaini2016anisotropicdesc}
\BIBentryALTinterwordspacing
D.~Boscaini, J.~Masci, E.~Rodolà, M.~M. Bronstein, and D.~Cremers,
  ``Anisotropic diffusion descriptors,'' \emph{Computer Graphics Forum},
  vol.~35, no.~2, pp. 431--441, 2016. [Online]. Available:
  \url{http://dx.doi.org/10.1111/cgf.12844}
\BIBentrySTDinterwordspacing

\bibitem{boscaini2016anisotropiccnn}
D.~Boscaini, J.~Masci, E.~Rodolà, and M.~Bronstein, ``Learning shape
  correspondence with anisotropic convolutional neural networks,'' in
  \emph{Advances in Neural Information Processing Systems 29}, Conference
  Proceedings.

\bibitem{monti2016geometric}
F.~Monti, D.~Boscaini, J.~Masci, E.~Rodol{\`a}, J.~Svoboda, and M.~M.
  Bronstein, ``Geometric deep learning on graphs and manifolds using mixture
  model cnns,'' \emph{arXiv preprint arXiv:1611.08402}, 2016.

\bibitem{bronstein2016geometricdeep}
\BIBentryALTinterwordspacing
M.~M. Bronstein, J.~Bruna, Y.~LeCun, A.~Szlam, and P.~Vandergheynst,
  ``Geometric deep learning: going beyond euclidean data,'' in \emph{ArXiv
  e-prints}, vol. 1611, Conference Proceedings. [Online]. Available:
  \url{http://adsabs.harvard.edu/abs/2016arXiv161108097B}
\BIBentrySTDinterwordspacing

\bibitem{GlorotAISTATS2010}
X.~Glorot and Y.~Bengio, ``Understanding the difficulty of training deep
  feedforward neural networks,'' in \emph{JMLR W\&CP: Proceedings of the
  Thirteenth International Conference on Artificial Intelligence and Statistics
  (AISTATS 2010)}, vol.~9, May 2010, pp. 249--256.

\bibitem{ioffe2015batch}
S.~Ioffe and C.~Szegedy, ``Batch normalization: Accelerating deep network
  training by reducing internal covariate shift,'' \emph{arXiv preprint
  arXiv:1502.03167}, 2015.

\bibitem{kingma2014adam}
D.~Kingma and J.~Ba, ``Adam: A method for stochastic optimization,''
  \emph{arXiv preprint arXiv:1412.6980}, 2014.

\bibitem{krizhevsky2009learning}
A.~Krizhevsky and G.~Hinton, ``Learning multiple layers of features from tiny
  images,'' 2009.

\bibitem{russakovsky2015imagenet}
O.~Russakovsky, J.~Deng, H.~Su, J.~Krause, S.~Satheesh, S.~Ma, Z.~Huang,
  A.~Karpathy, A.~Khosla, M.~Bernstein \emph{et~al.}, ``Imagenet large scale
  visual recognition challenge,'' \emph{International Journal of Computer
  Vision}, vol. 115, no.~3, pp. 211--252, 2015.

\bibitem{hcpOnline}
``Human connectome project,'' \url{https://db.humanconnectome.org}.

\bibitem{glasser2013mpp}
M.~F. Glasser \emph{et~al.}, ``The minimal preprocessing pipelines for the
  {H}uman {C}onnectome {P}roject,'' \emph{Neuroimage}, vol.~80, pp. 105--124,
  2013.

\bibitem{salimi2014adf}
G.~Salimi-Khorshidi, G.~Douaud, C.~F. Beckmann, M.~F. Glasser, L.~Griffanti,
  and S.~M. Smith, ``Automatic denoising of functional {MRI} data: combining
  independent component analysis and hierarchical fusion of classifiers,''
  \emph{Neuroimage}, vol.~90, pp. 449--468, 2014.

\bibitem{tzourio2002aal}
N.~Tzourio-Mazoyer, B.~Landeau, D.~Papathanassiou, F.~Crivello, O.~Etard,
  N.~Delcroix, B.~Mazoyer, and M.~Joliot, ``Automated anatomical labeling of
  activations in {SPM} using a macroscopic anatomical parcellation of the {MNI}
  {MRI} single-subject brain,'' \emph{Neuroimage}, vol.~15, no.~1, pp.
  273--289, 2002.

\bibitem{zhu2014constructing}
X.~Zhu, C.~Change~Loy, and S.~Gong, ``Constructing robust affinity graphs for
  spectral clustering,'' in \emph{Proc. CVPR}, 2014, pp. 1450--1457.

\bibitem{vergun2013characterizing}
S.~Vergun, A.~Deshpande, T.~B. Meier, J.~Song, D.~L. Tudorascu, V.~A. Nair,
  V.~Singh, B.~B. Biswal, M.~E. Meyerand, R.~M. Birn \emph{et~al.},
  ``Characterizing functional connectivity differences in aging adults using
  machine learning on resting state fmri data,'' \emph{Frontiers in
  computational neuroscience}, vol.~7, p.~38, 2013.

\bibitem{wale2008comparison}
N.~Wale, I.~A. Watson, and G.~Karypis, ``Comparison of descriptor spaces for
  chemical compound retrieval and classification,'' \emph{Knowledge and
  Information Systems}, vol.~14, no.~3, pp. 347--375, 2008.

\bibitem{yanardag2015deep}
P.~Yanardag and S.~Vishwanathan, ``Deep graph kernels,'' in \emph{Proceedings
  of the 21th ACM SIGKDD International Conference on Knowledge Discovery and
  Data Mining}.\hskip 1em plus 0.5em minus 0.4em\relax ACM, 2015, pp.
  1365--1374.

\bibitem{dobson2003distinguishing}
P.~D. Dobson and A.~J. Doig, ``Distinguishing enzyme structures from
  non-enzymes without alignments,'' \emph{Journal of molecular biology}, vol.
  330, no.~4, pp. 771--783, 2003.

\bibitem{shervashidze2009efficient}
N.~Shervashidze, S.~Vishwanathan, T.~Petri, K.~Mehlhorn, and K.~M. Borgwardt,
  ``Efficient graphlet kernels for large graph comparison.'' in \emph{AISTATS},
  vol.~5, 2009, pp. 488--495.

\bibitem{shervashidze2011weisfeiler}
N.~Shervashidze, P.~Schweitzer, E.~J.~v. Leeuwen, K.~Mehlhorn, and K.~M.
  Borgwardt, ``Weisfeiler-lehman graph kernels,'' \emph{Journal of Machine
  Learning Research}, vol.~12, no. Sep, pp. 2539--2561, 2011.

\bibitem{bosphorus3d}
\BIBentryALTinterwordspacing
A.~Savran, N.~Aly\"{u}z, H.~Dibeklio\u{g}lu, O.~\c{C}eliktutan, B.~G\"{o}kberk,
  B.~Sankur, and L.~Akarun, ``Biometrics and identity management,''
  B.~Schouten, N.~C. Juul, A.~Drygajlo, and M.~Tistarelli, Eds.\hskip 1em plus
  0.5em minus 0.4em\relax Berlin, Heidelberg: Springer-Verlag, 2008, ch.
  Bosphorus Database for 3D Face Analysis, pp. 47--56. [Online]. Available:
  \url{$http://dx.doi.org/10.1007/978-3-540-89991-4_6$}
\BIBentrySTDinterwordspacing

\bibitem{largeobjectset}
S.~Choi, Q.-Y. Zhou, S.~Miller, and V.~Koltun, ``A large dataset of object
  scans,'' \emph{arXiv:1602.02481}, 2016.

\bibitem{meshlab}
\BIBentryALTinterwordspacing
P.~Cignoni, M.~Corsini, and G.~Ranzuglia, ``Meshlab: an open-source 3d mesh
  processing system,'' \emph{ERCIM News}, no.~73, pp. 45--46, April 2008.
  [Online]. Available: \url{http://vcg.isti.cnr.it/Publications/2008/CCR08}
\BIBentrySTDinterwordspacing

\bibitem{NYUv2}
P.~K. Nathan~Silberman, Derek~Hoiem and R.~Fergus, ``Indoor segmentation and
  support inference from rgbd images,'' in \emph{ECCV}, 2012.

\bibitem{B3DO}
A.~Janoch, S.~Karayev, J.~Yangqing, J.~T. Barron, M.~Fritz, K.~Saenko, and
  T.~Darrell, ``A category-level 3-d object dataset: Putting the kinect to
  work,'' in \emph{2011 IEEE International Conference on Computer Vision
  Workshops (ICCV Workshops)}, Conference Proceedings, pp. 1168--1174.

\bibitem{RGBD}
K.~Lai, L.~Bo, X.~Ren, and D.~Fox, ``A large-scale hierarchical multi-view
  rgb-d object dataset,'' in \emph{IEEE International Conference on Robotics
  and Automation (ICRA)}, Conference Proceedings.

\bibitem{niyogi2004locality}
X.~He and P.~Niyogi, ``Locality preserving projections,'' in \emph{Neural
  information processing systems}, vol.~16.\hskip 1em plus 0.5em minus
  0.4em\relax MIT, 2004, p. 153.

\bibitem{deeplab}
\BIBentryALTinterwordspacing
L.~Chen, G.~Papandreou, I.~Kokkinos, K.~Murphy, and A.~L. Yuille, ``Deeplab:
  Semantic image segmentation with deep convolutional nets, atrous convolution,
  and fully connected crfs,'' \emph{CoRR}, vol. abs/1606.00915, 2016. [Online].
  Available: \url{http://arxiv.org/abs/1606.00915}
\BIBentrySTDinterwordspacing

\bibitem{docclassdata}
P.~Sen, G.~M. Namata, M.~Bilgic, L.~Getoor, B.~Gallagher, and T.~Eliassi-Rad,
  ``Collective classification in network data,'' \emph{AI Magazine}, vol.~29,
  no.~3, pp. 93--106, 2008.

\bibitem{yang2016revisiting}
Z.~Yang, W.~Cohen, and R.~Salakhutdinov, ``Revisiting semi-supervised learning
  with graph embeddings,'' \emph{arXiv preprint arXiv:1603.08861}, 2016.

\end{thebibliography}

%
\vspace{-12 mm}
\begin{IEEEbiography}[{\includegraphics[width=1in,height=1.25in,clip,keepaspectratio]{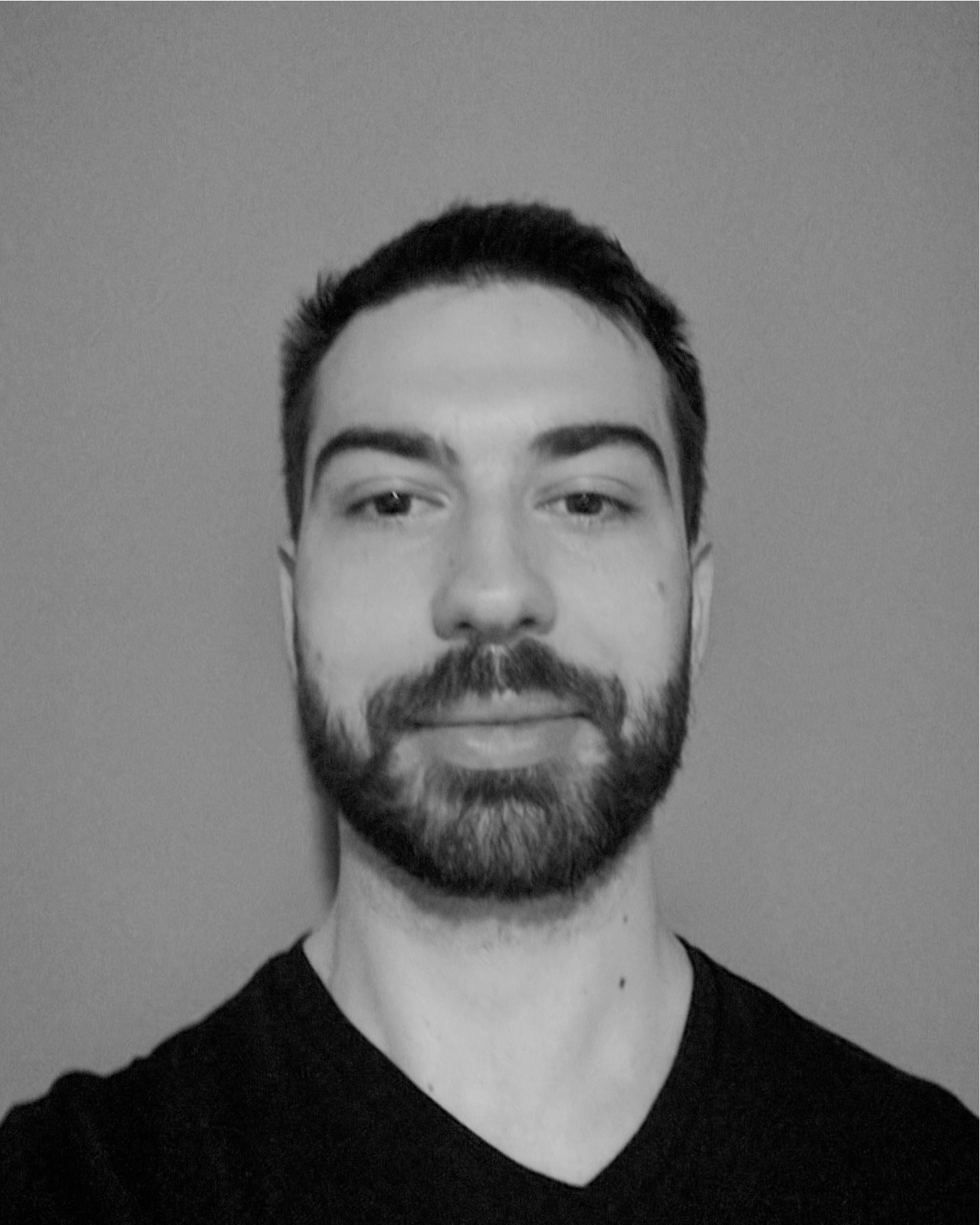}}]{Felipe Petroski Such} received his BS and MS in Computer Engineering at the Rochester Institute of Technology, New York, USA, in 2017. His interest in machine intelligence ranges from hardware acceleration to applicable software. During his time at RIT he worked as a Teaching Assistant for deep learning as well as a Research Assistant where he developed state-of-the-art handwriting recognition and graph filtering software.
\end{IEEEbiography}
\vspace{-12 mm}
\begin{IEEEbiography}[{\includegraphics[width=1in,height=1.25in,clip,keepaspectratio]{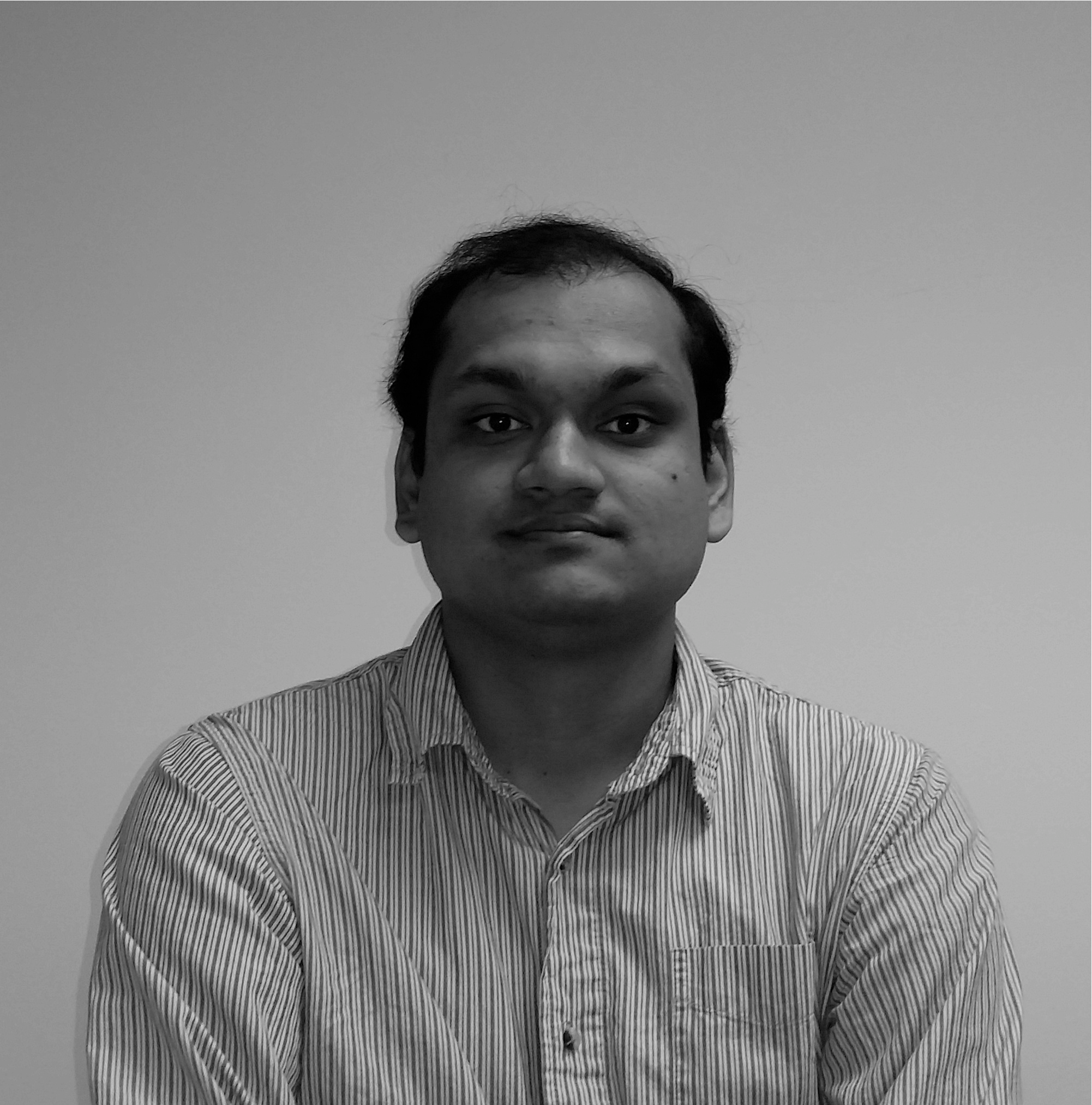}}]{Shagan Sah} obtained a Bachelors in Engineering degree from University of Pune, India and a M.S. in Imaging Science from Rochester Institute of Technology (RIT), USA with the aid of an RIT Graduate Scholarship. He is currently a Ph.D. candidate in the Center for Imaging Science at RIT. His current work and interests lie in the intersection of machine learning, natural language processing and computer vision applications in image and video understanding. In the past, he has worked at Motorola, Xerox-PARC and Cisco Systems.
\end{IEEEbiography}
\vspace{-12 mm}
\begin{IEEEbiography}[{\includegraphics[width=1in,height=1.25in,clip,keepaspectratio]{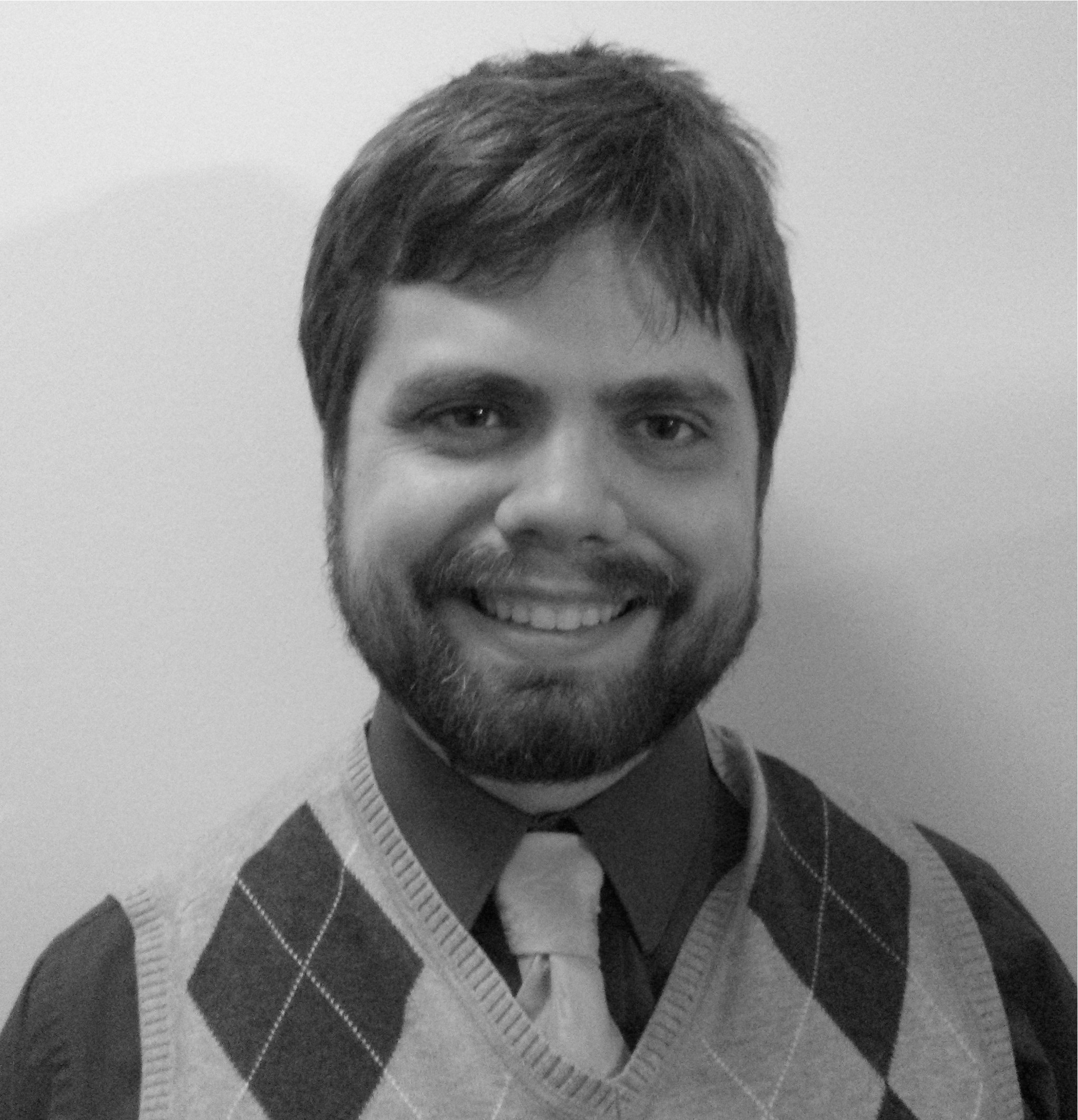}}]{Miguel Dominguez} earned a B.S. in Computer Science and Engineering from University of Toledo in 2012 and an M.S. in Electrical Engineering from Rochester Institute of Technology (RIT) in 2016. He is currently a GEM Fellow pursuing a Ph.D. in Engineering at RIT. His research interests include deep learning, computer vision, graph signal processing, and audio and speech processing. He is inspired to do research by his Blessed Mother.
\end{IEEEbiography}
\vspace{-12 mm}
\begin{IEEEbiography}[{\includegraphics[width=1in,height=1.25in,clip,keepaspectratio]{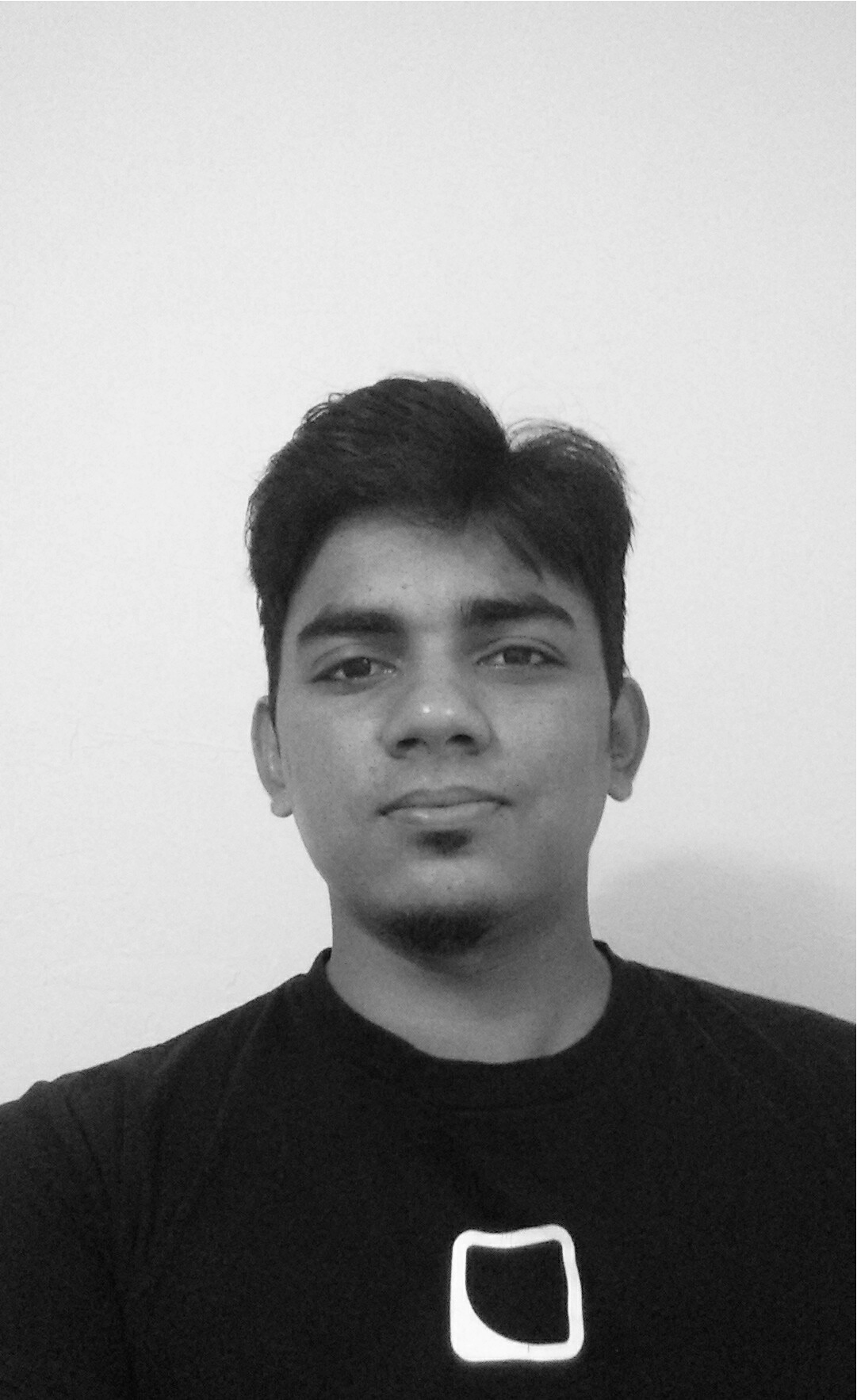}}]{Suhas Pillai} received his M.Sc in Computer Science from Rochester Institute of Technology in 2017. Since then he has joined TeraDeep as a Machine Learning Engineer, developing algorithms for detection of objects in satellite imagery. His research interests include speech and pattern recognition and computer vision.
\end{IEEEbiography}
\vspace{-12 mm}
\begin{IEEEbiography}[{\includegraphics[width=1in,height=1.25in,clip,keepaspectratio]{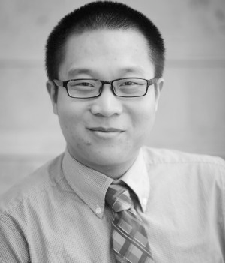}}]{Chao Zhang} is an imaging science student at Rochester Institute of Technology and Autism Developmental and Medicine Institute of Geisinger Health System. His research focus is on functional magnetic resonance image (fMRI), applying data mining to associate fMRI metrics with subject measures.
\end{IEEEbiography}
\vspace{-12 mm}
\begin{IEEEbiography}[{\includegraphics[width=1in,height=1.25in,clip,keepaspectratio]{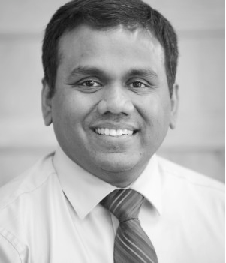}}]{Andrew Michael}, PhD received his BE in Electronics and Communication Engineering (First Class) in 2001 at the National Institute of Technology, India. He received his MS in Electrical Engineering and PhD in Imaging science at the Rochester Institute of Technology, New York, USA, in 2004 and 2009, respectively. In 2009 he joined the Mind Research Network as research scientist and program manager where he researched on data fusion methods to identify brain markers of psychiatric illnesses. In 2013 he joined Geisinger Health Systems in Pennsylvania as Assistant Professor and the founding director of the Neuroimaging Analytics Laboratory. At Geisinger, Dr. Michael contributes his signal processing, image analytics and machine learning skills to the analyses of brain imaging data from multiple imaging modalities.
\end{IEEEbiography}
\vspace{-12 mm}
\begin{IEEEbiography}[{\includegraphics[width=1in,height=1.25in,clip,keepaspectratio]{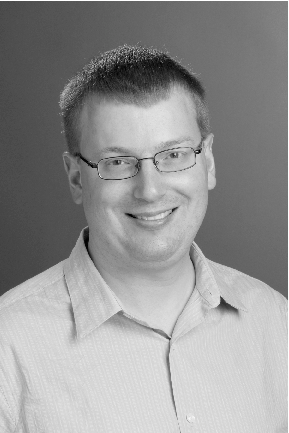}}]{Nathan D. Cahill} is the Associate Dean for Industrial Partnerships in the College of Science at RIT, where he is also an Associate Professor of Mathematical Sciences, a Graduate Program Faculty member of the Center for Imaging Science, and Director of the Image Computing and Analysis Laboratory. Prior to joining the RIT faculty in 2009, he spent 13 years in the Kodak Research Labs and at Carestream Health, during which time he was an inventor on 26 granted U.S. Patents and earned a DPhil in Engineering Science from the University of Oxford. His research is on the development of mathematical models and computational algorithms for solving problems in computer vision, medical imaging, and remote sensing.
\end{IEEEbiography}
\vspace{-12 mm}
\begin{IEEEbiography}[{\includegraphics[width=1in,height=1.25in,clip,keepaspectratio]{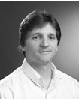}}]{Raymond Ptucha} is an Assistant Professor in Computer Engineering and Director of the Machine Intelligence Laboratory at Rochester Institute of Technology.  His research specializes in machine learning, computer vision, and robotics.  Ray was a research scientist with Eastman Kodak Company where he worked on computational imaging algorithms and was awarded 31 U.S. patents with another 19 applications on file. He graduated from SUNY/Buffalo with a B.S. in Computer Science and a B.S. in Electrical Engineering. He earned a M.S. in Image Science from RIT. He earned a Ph.D. in Computer Science from RIT in 2013. Ray was awarded an NSF Graduate Research Fellowship in 2010 and his Ph.D. research earned the 2014 Best RIT Doctoral Dissertation Award.  Ray is a passionate supporter of STEM education and is an active member of his local IEEE chapter and FIRST robotics organizations.
\end{IEEEbiography}





\end{document}